\documentclass[11pt]{article}

\usepackage{jmlr2e4arXiv}

\usepackage{amsfonts,amsmath,amssymb,mathtools}
\usepackage[pdfstartview={FitH},bookmarks=true,colorlinks=true,linkcolor=blue,citecolor=blue]{hyperref}
\usepackage{hyphenat,footmisc,float,geometry}
\usepackage{setspace}
\usepackage{lineno}

\newcommand{\diag}{{\rm diag}}
\newcommand{\rank}{{\rm rank}}

\newcommand{\real}{\mathbb{R}}

\newcommand{\iid}{\stackrel{i.i.d.}{\sim}}

\newcommand{\logit}{{\rm logit}}

\jmlrheading{00}{2015}{00-00}{00/00}{00/00}{Xiongzhi Chen and John D. Storey}

\ShortHeadings{Consistent Estimation of Latent Structure}{Chen and Storey}
\firstpageno{1}

\begin{document}

\title{Consistent Estimation of Low-Dimensional Latent Structure \\ in High-Dimensional Data}

\author{\name Xiongzhi Chen \email xiongzhi@princeton.edu \\
\name John D.\ Storey \email jstorey@princeton.edu \\
\addr Center for Statistics and Machine Learning and \\
Lewis-Sigler Institute for Integrative Genomics \\
Princeton University\\
Princeton, NJ 08544, USA \\
\ \\
{\rm October 2015}
}

\editor{ }

\maketitle

\begin{abstract}
We consider the problem of extracting a low-dimensional, linear latent variable structure from high-dimensional random variables. Specifically, we show that under mild conditions and when this structure manifests itself as a linear space that spans the conditional means, it is possible to consistently recover the structure using only information up to the second moments of these random variables. This finding, specialized to one-parameter exponential families whose variance function is quadratic in their means, allows for the derivation of an explicit estimator of such latent structure. This approach serves as a latent variable model estimator and as a tool for dimension reduction for a high-dimensional matrix of data composed of many related variables. Our theoretical results are verified by simulation studies and an application to genomic data.
\end{abstract}


\section{Introduction\label{Sec:Intro}}

Low-dimensional latent variable models are often used to capture systematic structure in the conditional mean space of high-dimensional random variables (rv's). This has been a popular strategy in high-dimensional probabilistic modeling and data analysis, and it serves as an attractive strategy for dimension reduction and recovering latent structure.  Examples include factor analysis \citep{BKM2011}, probabilistic principal components analysis (PCA) \citep{RSSB:RSSB196}, non-negative matrix factorization \citep{Lee1999}, asymptotic PCA \citep{Leek:2011}, latent Dirichlet allocation (LDA) \citep{Blei:2003}, and exponential family distribution extensions of PCA \citep{collins2001generalization}.

Let $\mathbf{Y}_{k\times n}=\left(  y_{ij}\right)  $ be an observed data matrix of $k$ variables (one variable per row), each with $n$ observations, whose entries
are rv's such that%
\begin{equation}
\theta_{ij}=\mathbb{E}\left[  y_{ij}|\mathbf{M}\right]  =\left(  \boldsymbol{\Phi
}\mathbf{M}\right)  \left(  i,j\right)  \text{ or }\boldsymbol{\Theta
}=\mathbb{E}\left[  \mathbf{Y}|\mathbf{M}\right]  =\boldsymbol{\Phi}\mathbf{M},
\label{7}%
\end{equation}
where $\mathbb{E}$  is the expectation operator, $\mathbf{M}_{r \times n}$ is a matrix of $r$ latent variables, and $\boldsymbol{\Phi}_{k \times r}$ is a matrix of coefficients relating the latent variables to the observed variables.  Furthermore, the dimensions are such as $k \gg n \geq r$.  In model (\ref{7}), the conditional mean $\theta_{ij}$ of $y_{ij}$ for a fixed $i$ only depends on the $j$th column $\mathbf{m}^{j}$ of $\mathbf{M}$, and
each row of the conditional mean matrix $\boldsymbol{\Theta}$ lies in the row space $\Pi_{\mathbf{M}}$ of $\mathbf{M}$. The latent structure in $\boldsymbol{\Theta}$ is therefore induced by $\Pi_{\mathbf{M}}$.

The above model is a general form of several highly used models.  This includes instances of factor analysis \citep{BKM2011}, probabilistic PCA \citep{RSSB:RSSB196}, mixed membership clustering in population genetics \citep{PritchardStephens2000,Alexander:2009p2792} which is closely related to LDA, and non-negative matrix factorization \citep{Lee1999}.  Whereas the specialized models are often focused on the probabilistic interpretation of the columns of $\mathbf{M}$, we are instead here interested in its row space, $\Pi_{\mathbf{M}}$.  This row space is sufficient for: (i) characterizing systematic patterns of variation in the data $\mathbf{Y}$, which can be used for exploratory data analysis or dimension reduction; (ii) accounting for the latent variables in downstream modeling of $\mathbf{Y}$ \citep{Leek:2007,Leek:2008} that requires adjustment for these variables; (iii) potentially identifying suitable initial values or geometric constraints for algorithms that estimate probabilistically constrained versions of $\mathbf{M}$; (iv) or recovering $\mathbf{M}$ itself if additional geometric properties are known (e.g., as in \cite{arora2013practical}).  Furthermore, many of the above models make assumptions about the probability distribution of $\mathbf{M}$ that may be untrue on a given data set.  One can compare our estimate of $\Pi_{\mathbf{M}}$ (which makes minimal assumptions) to the space induced by the model-based estimates of $\mathbf{M}$ to gauge the accuracy of the model assumptions and fit.  Therefore, we focus on estimation of the latent variable space $\Pi_{\mathbf{M}}$. Estimation of $\boldsymbol{\Phi}$ may be tractable, but we do not focus on that here.

\cite{Leek:2011} and \cite{Anandkumar:2012,Anandkumar:2015} have carried out work that is complementary to that presented here.  They both study moment estimators of linear latent variable models applied to high-dimensional data.   We explain how our work is related to \cite{Leek:2011} in \autoref{sec:Leek} and how it is related to \cite{Anandkumar:2012,Anandkumar:2015} in \autoref{sec:Anandkumar}.  The strategies employed in these papers have ties to what we do here; however, they each consider different probabilistic models with theory that does not directly apply to the models we study.

We show that both the row space $\Pi_{\mathbf{M}}$ of $\mathbf{M}$ (in \autoref{sec:estM}) and the row rank of $\mathbf{M}$ (in \autoref{sec:estr}) can be consistently estimated using information from a suitably adjusted $n \times n$ matrix $k^{-1} \mathbf{Y}^{T}\mathbf{Y}$.
In \autoref{Sec:AppToExpFamily}, we specialize these general results to $y_{ij}$'s that, conditional on $\mathbf{M}$, come from exponential family distributions. In particular, we explicitly construct a nonparametric, consistent estimator of the row space $\Pi_{\mathbf{M}}$ of $\mathbf{M}$ for $y_{ij}$ rv's that follow the natural exponential family (NEF) with quadratic variance function (QVF) using information only up to their second moments, and the estimator is computationally straightforward to implement. In \autoref{sec:ranPhi}, we extend the results of previous sections to the case where $\boldsymbol{\Phi}$ is random.  A simulation study is conducted in \autoref{Sec:SimThyEstM} to check and confirm our theoretical findings, and we apply the estimators to a genomics data set in \autoref{sec:rnaseq}. Finally, we end the article with a discussion in \autoref{Sec:Disc}, collect in \autoref{AppB:proof} all technical proofs, and present the the full set of results from simulation studies in \autoref{AppC:ResSimu} and \autoref{AppD:ResSimu}.

\section{Almost Surely Estimating the Latent Linear Space \label{Sec:NPEstM} \label{sec:estM}}
The big picture strategy we take is summarized as follows.  Carrying out a decomposition related to \cite{Leek:2011}, we first expand $k^{-1}\mathbf{Y}^{T}\mathbf{Y}$ into four components:
\begin{eqnarray}
k^{-1}\mathbf{Y}^{T}\mathbf{Y} & = & k^{-1}(\mathbf{Y} - \boldsymbol{\Phi} \mathbf{M} + \boldsymbol{\Phi} \mathbf{M})^T (\mathbf{Y} - \boldsymbol{\Phi} \mathbf{M} + \boldsymbol{\Phi} \mathbf{M}) \nonumber \\
\ & = & \underbrace{k^{-1}(\mathbf{Y} - \boldsymbol{\Phi} \mathbf{M})^T (\mathbf{Y} - \boldsymbol{\Phi} \mathbf{M})}_\text{$(i)$} + \underbrace{k^{-1}(\mathbf{Y} - \boldsymbol{\Phi} \mathbf{M})^T (\boldsymbol{\Phi} \mathbf{M})}_\text{$(ii)$} \\
\ & \ & + \underbrace{k^{-1}(\boldsymbol{\Phi} \mathbf{M})^T (\mathbf{Y} - \boldsymbol{\Phi} \mathbf{M})}_\text{$(iii)$} + \underbrace{k^{-1}(\boldsymbol{\Phi} \mathbf{M})^T (\boldsymbol{\Phi} \mathbf{M})}_\text{$(iv)$} \label{eq:terms}
\end{eqnarray}
We then show the following about each of the components as $k \rightarrow \infty$, under the assumptions given in detail below:
\begin{itemize}
\item $(i)$: this term may be estimated arbitrarily well as $k \rightarrow \infty$ by a diagonal matrix defined and studied below;
\item $(ii)$ and $(iii)$: these terms converge to the zero matrix;
\item $(iv)$: this term converges to a symmetric matrix with $r$ positive eigenvalues whose leading $r$ eigenvectors span the row space of $\mathbf{M}$.
\end{itemize}
Once the convergence of these terms is rigorously established, the strategy we take is to form an estimate of term $(i)$, denoted by $\mathbf{\hat{D}}_{k}$, and show that the space spanned by the leading $r$ eigenvectors of $k^{-1}\mathbf{Y}^{T}\mathbf{Y} - \mathbf{\hat{D}}_{k}$ converges to the row space of $\mathbf{M}$ as $k \rightarrow \infty$.  We then also provide a framework to estimate the dimension $r$ of the row space of $\mathbf{M}$ and incorporate it into this estimation framework.

\subsection{Model Assumptions}
We first define the matrix norm $\left\Vert \mathbf{X}\right\Vert =\sqrt{\sum\limits_{i,j}x_{ij}^{2}}$ for any
real matrix $\mathbf{X}=\left(  x_{ij}\right)  $ and let $C>0$ denote a generic, finite
constant whose value may vary at different occurrences. We first assume that $\mathbf{\Phi}$ is deterministic; the results in this section are extended to the case where $\mathbf{\Phi}$ is random in \autoref{sec:ranPhi}. The assumptions on
model (\ref{7}) are as follows:

\begin{description}
\item[A1)] $1 \leq \rank\left(  \mathbf{M}\right)  =r<n$ and $n$ is finite; $\left\{  y_{ij} | \mathbf{M}\right\}
_{i,j}$ are jointly independent with variance $\mathbb{V}\left[
y_{ij}|\mathbf{M}\right]  =\delta_{ij}$ such that $\sup\limits_{k}\max\limits_{i,j}\mathbb{E}
\left[  y_{ij}^{8}|\mathbf{M}\right]  \leq C$ (which implies $\sup\limits_{k}\max\limits
_{i,j}\mathbb{E}\left[  y_{ij}^{4}|\mathbf{M}\right]  \leq C$), where $\mathbb{V}$
is the variance operator.

\item[A2)] $\sup\limits_{k \geq 1}\max\limits_{1\leq i \leq k}\left\Vert \boldsymbol{\phi}_{i}\right\Vert \leq C$, where $\boldsymbol{\phi}_{i}$ is the $i$th row of $\boldsymbol{\Phi}$. Further, for some $\mathbf{W}_{r\times r}>0$ and some non-negative sequence
$c_{k}\rightarrow0$,
\begin{equation}
\left\Vert k^{-1}\boldsymbol{\Phi}^{T}\boldsymbol{\Phi}-\mathbf{W}\right\Vert
=c_{k}\text{.}\label{2}
\end{equation}
\end{description}
Since we are considering model (\ref{7}) for which $\mathbf{Y}$ is conditioned on $\mathbf{M}$,
all random vectors in this model are by default conditioned on $\mathbf{M}$ unless otherwise noted (e.g., \autoref{sec:ranPhi}). For conciseness we will omit stating ``conditional on $\mathbf{M}$'' in this default setting.

 We state a consequence of the assumption A2) as Lemma \ref{Lm:UnifConv}, whose proof is straightforward and omitted.

\begin{lemma}
\label{Lm:UnifConv}If the assumption A2) holds, then
\[
\lim_{k\rightarrow\infty}k^{-1}\sum_{i=1}^{k}\theta_{ij}^{2}=\left(  \mathbf{M}%
^{T}\mathbf{WM}\right)_{jj}
\]
for each $1\leq j\leq n$ and
$\sup\limits_{k}\max\limits_{i,j}\left\vert \theta_{ij}\right\vert \leq C$.
\end{lemma}
The uniform boundedness results provided by Lemma \ref{Lm:UnifConv} will be used to prove the convergence results later.

\subsection{$k^{-1}\mathbf{Y}^{T}\mathbf{Y}$ Asymptotically Preserves the Latent Linear Space}

We first derive the
asymptotic form of $k^{-1}\mathbf{Y}^{T}\mathbf{Y}$ with the aid of
the strong law of large numbers (SLLN) in \cite{Walk:2005}.  Let $\bar{\delta}_{kj}=k^{-1}\sum\limits_{l=1}^{k}\delta_{lj}$ be the column-wise average variance of $\mathbf{Y} | \mathbf{M}$ (where $\mathbb{V}\left[y_{ij}|\mathbf{M}\right]  =\delta_{ij}$ as defined above) and
\begin{equation}
\mathbf{D}_{k}=\mbox{diag}\left\{  \bar{\delta}_{k1},...,\bar{\delta}_{kn}\right\}\label{eq:Dk}
\end{equation}
the $n \times n$ diagonal matrix composed of these average variances.

\begin{theorem}
\label{Thm:AsympReducBzero}Under the assumptions for model (\ref{7}),
\begin{equation}
\lim_{k\rightarrow\infty}\left\Vert k^{-1}\mathbf{Y}^{T}\mathbf{Y}-\mathbf{D}_{k}
-\mathbf{H}  \right\Vert =0 \text{\ almost surely (a.s.),}\label{6}%
\end{equation}
where $\mathbf{H}=\mathbf{M}^{T}\mathbf{WM}$.
\end{theorem}

\autoref{Thm:AsympReducBzero} shows that $\mathbf{R}_{k}=k^{-1}%
\mathbf{Y}^{T}\mathbf{Y}-\mathbf{D}_{k}$ becomes arbitrarily close to
$\mathbf{H}$ as the number of variables $k \rightarrow \infty$. In fact, it gives much more information
on the eigensystem of $\mathbf{R}_{k}$ (as we take the convention that an
eigenvector always has norm $1$). Let $\left\{  \beta_{k,i}\right\}
_{i=1}^{n}$ be the eigenvalues of $\mathbf{R}_{k}$ ordered into $\beta
_{k,i}\geq\beta_{k,i+1}$ for $1\leq i\leq n-1$ (where here we take the convention
that the ordering of the designated multiple copies of a multiple eigenvalue
is arbitrary),
\[
S={\bigcup}_{i=1}^{r}\left\{  \mathbf{u}:\mathbf{u}\text{ is
an eigenvector of }\mathbf{R}_{k}\text{ corresponding to }\beta_{k,i}\right\}
\text{,}
\]
and $\left\{  \alpha_{i}\right\}  _{i=1}^{n}$ be
the eigenvalues of $\mathbf{H}$ ordered into $\alpha_{i}\geq\alpha_{i+1}$ for $1\leq i\leq n-1$.

\begin{corollary}
\label{Lm:ConvEigensys}Under the assumptions for model (\ref{7}),
\begin{equation}
\lim_{k\rightarrow\infty}\max_{1\leq i\leq n}\left\vert \beta_{k,i}-\alpha_{i}\right\vert =0\text{ a.s.}\label{1a}
\end{equation}
Further, $\lim\limits_{k\rightarrow\infty}\left\vert S\right\vert =r$ a.s.\ and%
\begin{equation}
\lim_{k\rightarrow\infty}\left\langle \left\{  \mathbf{u}\in S\right\}
\right\rangle \triangle\Pi_{\mathbf{M}}=\varnothing\text{ a.s.,}\label{3a}%
\end{equation}
where $\left\langle \cdot\right\rangle$ denotes the linear
space spanned by its arguments, and $\triangle$ the symmetric set difference.
\end{corollary}

Corollary \ref{Lm:ConvEigensys} reveals that asymptotically as $k\rightarrow\infty$ the eigenvalues of $\mathbf{R}_k$ converge to those of $\mathbf{H}$ when both sets of eigenvalues are ordered the same way, that the dimension of the space spanned by all the eigenvectors
corresponding to the $r$ largest eigenvalues of $\mathbf{R}_k$ converges to $r$ as $k \rightarrow \infty$, and that $\Pi_{\mathbf{M}}$ is asymptotically spanned by the leading $r$ dimensional joint eigenspace induced by
$\mathbf{R}_{k}$ as $k\rightarrow\infty$. When the nonzero eigenvalues of
$\mathbf{H}$ are distinct, we easily have%
\begin{equation}
\lim_{k\rightarrow\infty}\left\langle \mathbf{u}_{k,i}\right\rangle
\triangle\left\langle \mathbf{v}_{i}\right\rangle =\varnothing \text{ a.s. for each }
 i=1,\ldots,r,\label{2a}%
\end{equation}
where, modulo a sign, $\mathbf{u}_{k,i}$ is the eigenvector corresponding to
$\beta_{k,i}$ and $\mathbf{v}_{i}$ that to $\alpha_{i}$.

When the dimension $r$ of the latent space $\Pi_{\mathbf{M}}$ and the diagonal matrix $\mathbf{D}_{k}$ of the column-wise average variances are known, it follows by Corollary \ref{Lm:ConvEigensys} that $\tilde{\Pi}_{\mathbf{M}}=\left\langle \left\{  \mathbf{u}\in S\right\}  \right\rangle $ asymptotically spans the latent space $\Pi_{\mathbf{M}}$, and $\tilde{\Pi}_{\mathbf{M}}$ converges to the row space $\Pi_{\mathbf{M}}$ with probability 1. However, in practice both $r$ and $\mathbf{D}_{k}$ need to be estimated, which is the topic of the next three sections. Estimating the number of latent variables $r$ is in general a difficult problem.  In our setting we also must accurately estimate $\mathbf{D}_{k}$, which can be a difficult task when the variances $\delta_{ij}$ may all be different (i.e., heteroskedastic).

\section{Consistently Estimating the Latent Linear Space Dimension} \label{sec:estr}
The strategy we take to consistently estimate the dimension of the latent variable space $r$ is to carefully scale the ordered eigenvalues of $\mathbf{R}_{k}=k^{-1}\mathbf{Y}^{T}\mathbf{Y}-\mathbf{D}_{k}$ and identify the index of the eigenvalue whose magnitude separates the magnitudes of these eigenvalues into two particular groups when $k$ is large. Recall that, by \autoref{Thm:AsympReducBzero} and Corollary \ref{Lm:ConvEigensys}, the difference between the vector of descendingly ordered eigenvalues of $\mathbf{R}_{k}$ and that of those of $\mathbf{H}=\mathbf{M}^{T}\mathbf{WM}$ converges to zero as $k \rightarrow \infty$. However, since $\rank\left(\mathbf{H}\right) = r$ and $\mathbf{W} >0$, we know that the $r$ largest eigenvalues of $\mathbf{H}$ are positive but the rest are zero. This means that the $r$ largest eigenvalues of $\mathbf{R}_{k}$ are all strictly positive as $k \rightarrow \infty$, while the smallest $n-r$ eigenvalues of $\mathbf{R}_{k}$ converge to $0$ as $k \rightarrow \infty$. Depending on the speed of convergence of the $n-r$ smallest eigenvalues of $\mathbf{R}_{k}$ to $0$, if we suitably scale the eigenvalues of $\mathbf{R}_{k}$, then the scaled, ordered eigenvalues will eventually separate into two groups, those with very large magnitudes and the rest very small. The index of the scaled, ordered eigenvalues for which such a separation happens is then a consistent estimator
of $r$. If we replace $\mathbf{R}_{k}$ with an estimator $\hat{\mathbf{R}}_{k}=k^{-1}\mathbf{Y}^{T}\mathbf{Y}- \hat{\mathbf{D}}_{k}$ that satisfies a certain level of accuracy detailed below, then the previous reasoning applied to the eigenvalues of $\hat{\mathbf{R}}_{k}$ will also give a consistent estimator of $r$.

To find the scaling sequence for the magnitudes of the eigenvalues of $\mathbf{R}_{k}$ or equivalently the speed of convergence of
the $n-r$ smallest eigenvalues of $\mathbf{R}_{k}$ to $0$, we define the $k \times n $ matrix $\mathbf{E}$ with entry $\mathbf{E}(i,j)=e_{ij}=y_{ij}-\theta_{ij}
$, and study as a whole the random part of $k^{-1}\mathbf{Y}^{T}\mathbf{Y}$ defined by
\begin{equation}
\mathbf{F}=
\underbrace{k^{-1}\mathbf{E}^{T}\mathbf{E}}_{(i)} +
\underbrace{k^{-1}\mathbf{E}^{T}\boldsymbol{\Theta}}_{(ii)}  +
\underbrace{k^{-1}\boldsymbol{\Theta}^{T}\mathbf{E}}_{(iii)}
\label{22}.
\end{equation}
Note that the terms $(i)$, $(ii)$ and $(iii)$ in $\mathbf{F}$ correspond to those from equation \eqref{eq:terms}.
We will show that $\mathbf{F}$ configured as a $2n^2$ vector possesses asymptotic Normality after centering and scaling as $k \rightarrow \infty$. This then reveals that the scaling sequence for the eigenvalues of $\mathbf{R}_{k}$ should be no smaller than being proportional to
$k^{-1/2}$.

Let $\boldsymbol{\Delta}=\left(  \delta_{ij}\right)  $,
$\boldsymbol{\Delta}_{i}=\diag\left\{  \delta_{i1},...,\delta_{in}\right\}$ and define
\begin{equation}
\mathbf{z}_{i}=k^{-1}\left(  e_{i1}\mathbf{e}_{i},...,e_{in}\mathbf{e}%
_{i},\boldsymbol{\phi}_{i}\mathbf{m}^{1}\mathbf{e}_{i},...,\boldsymbol{\phi
}_{i}\mathbf{m}^{n}\mathbf{e}_{i}\right)  \label{24}%
\end{equation}
for $1\leq i\leq k$, where $\mathbf{e}_{i}$ and
$\mathbf{m}^{j}$ are respectively the $i$th row of $\mathbf{E}$ and the $j$th column of $\mathbf{M}$.
We have:
\begin{proposition}
\label{Prop:CLTRandomPart}  Under model \eqref{7}, $\mathbf{F}$ is a linear function of $\sum\limits_{i=1}^{k}\mathbf{z}_{i}$ with global Lipschitz constant $1$.  If we assume A1) and A2) from \autoref{Sec:NPEstM} and also
\begin{description}
\item[A3)] The sequences $k^{-1}\sum\limits_{i=1}^{k}\mathbb{E}\left[  e_{ij}%
^{4}|\mathbf{M}\right]  $, $k^{-1}\sum\limits_{i=1}^{k}\boldsymbol{\phi}_{i}\mathbb{E}\left[
e_{ij}^{3}|\mathbf{M}\right]  $ and $k^{-1} \sum_{i=1}^{k} \boldsymbol{\phi}_i \mathbf{m}^{j} \boldsymbol{\phi}_i \mathbf{m}^l\boldsymbol{\Delta}_i $ for any $1 \leq j \leq l \leq n$
are all convergent as $k \rightarrow \infty$,
\end{description}
then $\sum\limits_{i=1}^{k}\sqrt{k}\left(
\mathbf{z}_{i}-\mathbb{E}\left[  \mathbf{z}_{i}|\mathbf{M} \right]  \right)  $ converges
in distribution to a multivariate Normal random vector with mean zero.
\end{proposition}

With the concentration property of $\mathbf{F}$ established by
Proposition \ref{Prop:CLTRandomPart}, we will be able to explore the possibility of
consistently estimating $r$ by studying the magnitudes of eigenvalues
$\left\{  \hat{\alpha}_{k,i}\right\}  _{i=1}^{n}$ of
\begin{equation}
\mathbf{\hat{R}}_{k}=k^{-1}\mathbf{Y}^{T}\mathbf{Y}-\mathbf{\hat{D}}_{k}\text{
with }\mathbf{\hat{D}}_{k}=\diag\left\{  \hat{\delta}_{k1},...,\hat{\delta
}_{kn}\right\}  \text{,}\label{23}%
\end{equation}
where each $\hat{\delta}_{kj}$ is an estimate of $\bar{\delta}_{kj}$ for
$1\leq j\leq n$, and the $\hat{\alpha}_{k,i}$'s are
ordered into $\hat{\alpha}_{k,i}\geq\hat{\alpha}_{k,i+1}$ for $1\leq i<n-1$.

\begin{theorem}
\label{Prop:EstRank}Under the assumptions A1), A2) and A3), if
\begin{description}
\item [A4)] for some non-negative $\varepsilon_{k}$ such that $\varepsilon_{k}\rightarrow0$,
\begin{equation}
\left\Vert \mathbf{\hat{D}}_{k}-\mathbf{D}_{k}\right\Vert =\varepsilon_{k}\label{11},%
\end{equation}
\end{description}
then
\begin{equation}
\left\Vert \mathbf{\hat{R}}_{k}-\mathbf{H}\right\Vert =O_{\Pr}\left(  \tau
_{k}\right)  \text{ and }\left\Vert \hat{\alpha}_{k,i}-\alpha_{i}\right\Vert
=O_{\Pr}\left(  \tau_{k}\right)  \text{,}\label{26}%
\end{equation}
where $\Pr$ is the probability measure and
\begin{equation}
\tau_{k}=\max\left\{k^{-1/2},\varepsilon_{k},c_{k}\right\}
\text{.}\label{25}%
\end{equation}
Further, for any $\tilde{\tau}_{k}>0$ such that $\tilde{\tau}_{k}=o\left(
1\right)  $ and $ \tau_{k}=o\left( \tilde{\tau}_{k}\right)  $, as
$k\rightarrow\infty$,%
\begin{equation}
\left\{
\begin{tabular}
[c]{lll}%
$\Pr\left(  \tilde{\tau}_{k}^{-1}\hat{\alpha}_{k,i}>\tilde{c}\right)  \rightarrow0$ &
if & $r+1\leq i\leq n$,\\
$\Pr\left(  \tilde{\tau}_{k}^{-1}\hat{\alpha}_{k,i}\rightarrow\infty\right)
\rightarrow1$ & if & $1\leq i\leq r$,
\end{tabular}
\ \ \ \ \right.  \label{20}%
\end{equation}
for any fixed $\tilde{c}>0$. Therefore, letting
\begin{equation}\label{eq:rEstDef}
\hat{r} = \sum_{i=1}^{n} 1_{\left\{\tilde{\tau}_{k}^{-1}\hat{\alpha}_{k,i}>\tilde{c}\right\}}
\end{equation}
gives
\begin{equation}
\Pr\left(\hat{r}  = r \right) \rightarrow 1 \text{\ as \ }
k \rightarrow \infty, \label{eq:rankest}
\end{equation}
where $1_{A}$ is the indicator of a set $A$.
\end{theorem}

\autoref{Prop:EstRank} shows that the speed of convergence, $\left\{  \tau_{k}\right\}
_{k\geq1}$, of the eigenvalues of $\hat{\mathbf{R}}_k$ (and those of $\mathbf{R}_k$) is essentially is determined by those related to $k^{-1} \boldsymbol{\Phi}^{T}\boldsymbol{\Phi}$ and
$\mathbf{D}_{k}$; see equations \eqref{11}--\eqref{25}. Further, it reveals that, when $\left\{  \tau_{k}\right\}
_{k\geq1}$ is known, with probability approaching to $1$ as $k\rightarrow\infty$, the scaled, ordered eigenvalues $\left\{  \tilde{\tau}_{k}^{-1}\hat{\alpha}_{k,i}\right\}  _{i=1}^{n}$ eventually separates into
two groups: $\tilde{\tau}_{k}^{-1}\hat{\alpha}_{k,i}$ for
$1\leq i\leq r$ all lie above $\tilde{c}$ but the rest all lie below $\tilde{c}$ for any chosen $\tilde{c}>0$.
In other words, for a chosen $\tilde{c}>0$, the number of $\tilde{\tau}_{k}^{-1}\hat{\alpha}_{k,i} > \tilde{c}$ when $k$ is
large is very likely equal to $r$. However, in practice
$\left\{  \varepsilon_{k}\right\}  _{k\geq1}$ and $\left\{
c_{k}\right\}  _{k\geq1}$ are unknown, and even if they are known or can be estimated,
unfortunately the hidden constants in \eqref{26} are
unknown (even when $k=\infty$). Further, when $k$ is finite, the $n-r$ smallest
eigenvalues of $\mathbf{\hat{R}}_{k}$ may not yet be identically zero,
the hidden constants may have a large impact on estimating the scaling sequence $\tilde{\tau}_{k}$,
and rates slightly different than $\tilde{\tau}_{k}$ may have
to be chosen to balance the effects of the hidden constants;
see in \autoref{Sec:SimThyEstM} a brief discussion on the estimation of $\tilde{\tau}_{k}$ and
the effects of choosing $\tilde{\tau}_{k}$ on estimating $r$ for finite $k$.

Before we apply our theory to $y_{ij}$'s that follow specific parametric distributions, we pause to comment on the key results obtained so far. \autoref{Thm:AsympReducBzero} and Corollary \ref{Lm:ConvEigensys} together ensure that asymptotically as $k \rightarrow \infty$ we can span the row space $\Pi_{\mathbf{M}}$ of $\mathbf{M}$ by the $r$ leading eigenvectors of
$\mathbf{R}_k= k^{-1}\mathbf{Y}^{T}\mathbf{Y} - \mathbf{D}_k$ (see the definition of $\mathbf{D}_k$ in
\eqref{eq:Dk}). However, the conclusions in these results are based on a known $r$ and $\mathbf{D}_k$. In contrast, Proposition \ref{Prop:CLTRandomPart} and \autoref{Prop:EstRank} retain similar assertions to those of \autoref{Thm:AsympReducBzero} and Corollary \ref{Lm:ConvEigensys} by replacing the unknown $\mathbf{D}_k$ by its consistent estimate $\hat{\mathbf{D}}_k$, and they enable us to construct a consistent estimate $\hat{r}$ of $r$ such that the linear space spanned by the leading $\hat{r}$ eigenvectors of $\hat{\mathbf{R}}_k= k^{-1}\mathbf{Y}^{T}\mathbf{Y} - \hat{\mathbf{D}}_k$ consistently estimates $\Pi_{\mathbf{M}}$ as $k \rightarrow \infty$.
This reveals that to consistently estimate $\Pi_{\mathbf{M}}$ in a fully data driven approach using our theory, it is crucial to develop a consistent estimate $\hat{\mathbf{D}}_k$ of $\mathbf{D}_k$.

\section{Specializing to Exponential Family Distributions \label{Sec:AppToExpFamily}}

We specialize the general results obtained in \autoref{Sec:NPEstM} and \autoref{sec:estr} for model
(\ref{7}) to the case when $y_{ij}$ follow the single parameter exponential family probability density function (pdf) given
by
\begin{equation}
f(y;\theta)=h\left(  y\right)  \exp\left\{  \eta(\theta)y-g\left(  \eta\left(\theta\right)  \right)  \right\}, \label{1}
\end{equation}
where $\eta$ is the canonical link function, and $g$ and $h$ are known functions such that $f(y;\theta)$ is a proper pdf.  The values that $\theta$ can take are $\Omega=\left\{  \theta\in \mathbb{R}:\int f\left(  y;\theta\right)  =1\right\} $.
The following corollary says that our general results on consistent estimation of the latent space $\Pi_{\mathbf{M}}$
and its dimension $r$ hold when the link function $\eta$ is bounded on the closure of the parameter space $\Omega_0 \subseteq \Omega$.
\begin{corollary}
\label{Cor:GenToExp}Suppose $\Omega_0$ is an open
subset of $\Omega$ such that its closure lies in the interior of
$\Omega$. If conditional on $\mathbf{M}$ each $y_{ij}$ has pdf of the form (\ref{1})
with $\theta_{ij}\in\Omega_0$ and $\eta$ is bounded
as a function of $\theta$ in the closure of $\Omega_0$, then $\sup\limits_{k}\max\limits_{i,j}\mathbb{E}
\left[  y_{ij}^{8}|\mathbf{M}\right]  \leq C$, which implies that \autoref{Thm:AsympReducBzero},
Corollary \ref{Lm:ConvEigensys}, Proposition \ref{Prop:CLTRandomPart} and
\autoref{Prop:EstRank} hold.
\end{corollary}

We remark that the boundedness of $\eta$ on the closure of $\Omega_0$ is not restrictive, in that
in practice either $\Omega_0$ is bounded in Euclidean norm or $\eta$ is bounded in supremum norm.
The proof of Corollary \ref{Cor:GenToExp} is a simple observation that
$g$ is analytic in $\eta$ when $\theta \in \Omega_0$ (see, e.g. \citealp{Letac:1990}) and that
the derivative of $g$ in $\eta$ of any order is bounded when $\eta$ is bounded; the proof is thus omitted.

\subsection{Estimating $\mathbf{D}_k$}\label{subsec:apprAvgVar}

With Corollary \ref{Cor:GenToExp} (see also the discussion at the end of \autoref{sec:estr}), we only need to estimate $\bar{\delta}_{kj}$ (which in turn yields $\mathbf{D}_k$) in
order to consistently estimate $r$ and $\Pi_{\mathbf{M}}$. To obtain an estimate $\hat{\delta}_{kj}$ of $\bar{\delta}_{kj}$ when potentially all $\delta_{ij}$ are different from each other (i.e., complete heteroskedasticity), we exploit the intrinsic relationship between $\theta_{ij}$ and $\delta_{ij}$ when $y_{ij}$ come from a certain class of natural exponential family (NEF) distributions.

\begin{lemma}
\label{Lm:NEFQVF}Let $y$ have marginal pdf (\ref{1}). Then there exists a quadratic
function $v\left(  \cdot\right)  $ such that $\mathbb{E}\left[  v\left(
y\right)  \right]  =\mathbb{V}\left[  y\right]  $ if and only if $f$
parametrized by $\eta$ forms an NEF with quadratic variance function (QVF)
defined by \cite{Morris:1982} such that
\begin{equation}
\mathbb{V}\left[  y\right]  =b_{0}+b_{1}\mathbb{E}\left[  y\right]
+b_{2}\left(  \mathbb{E}\left[  y\right]  \right)  ^{2}\text{ with }b_{2}
\neq-1\label{28}
\end{equation}
for some $b_{0}, b_{1}, b_{2} \in\mathbb{R}$. Specifically, (\ref{28})
implies
\begin{equation}
v\left(  t\right)  =\left(  1+b_{2}\right)  ^{-1}\left(  b_{0}+b_{1} t+b_{2}t^{2}\right)  \text{.}\label{27}
\end{equation}
\end{lemma}

The proof of Lemma \ref{Lm:NEFQVF} is straightforward and omitted.
\autoref{tab:vfunc} lists $v\left(  \cdot\right)  $ for the six NEFs
with QVF.

\begin{table}[!t]
\centering
\caption{The function $v(\cdot)$ such that $\mathbb{E}[v(y)] = \mathbb{V}[y]$ when $y$ has pdf \eqref{1} and comes from a NEF with QVF in \cite{Morris:1982}. Note that $\logit(x)=\log{\dfrac{x}{1-x}}$ for $x \in (0,1)$ and ``GHS''
stands for ``generalized hyperbolic secant distribution''.}
\label{tab:vfunc}
\begin{tabular}{|l|c|c|c|c|}
\hline
\ &  $\theta$ & $\mathbb{V}[y]$ & $v(y)$ & $\eta(\theta)$ \\
\hline
Normal$(\mu,1)$  & $\mu$ & 1  & $1$ & $\theta$ \\				
Poisson$(\lambda)$ & $\lambda$ & $\theta$ &  $y$  & $\log(\theta)$ \\
Binomial$(s,p)$ & $sp$ & $\theta-\theta^2/s$ &  $(sy-y^2)/(s-1)$ & $\logit(\theta/s)$ \\
NegBin$(s,p)$ & $sp/(1-p)$ & $\theta+\theta^2/s$  &  $(sy + y^2)/(s+1)$ & $\log(\theta/(s+\theta))$ \\ 
Gamma$(s,\lambda)$ & $s/ \lambda$& $\theta^2/s$& $y^2/(1+s)$& $-1/\theta$\\
GHS$(s,\lambda)$ & $s \lambda$& $s+\theta^2/s$& $(s^2+y^2)/(1+s)$& $\arctan(\theta/s)$\\
\hline
\end{tabular}
\end{table}

Inspired by the availability of the function $v(\cdot)$ obtained in Lemma \ref{Lm:NEFQVF}, we now state a general result
on how to explicitly construct a $\hat{\mathbf{D}}_{k}$ that properly estimates $\mathbf{D}_{k}$.

\begin{lemma}
\label{Lm:AveVar}Let $y_{ij}$ have pdf (\ref{1}) such that $\mathbb{V}\left[
y_{ij}|\mathbf{M}\right]  =\mathbb{E}\left[  v\left(  y_{ij}\right) |\mathbf{M} \right]  $ for some
function $v\left(  \cdot\right)  $ satisfying%
\begin{equation}
\sup_{k}\max_{i,j}\mathbb{E}\left[  v^{4}\left(  y_{ij}\right)|\mathbf{M}
\right]  \leq C<\infty\label{29}.%
\end{equation}
Then
\begin{equation}
\hat{\delta}_{kj}=k^{-1}\sum\limits_{l=1}^{k}v\left(  y_{lj}\right)
\label{19}%
\end{equation}
satisfies $\lim\limits_{k\rightarrow\infty}\left\vert \hat{\delta}_{kj}-\bar{\delta
}_{kj}\right\vert =0$ a.s. for each $1\leq j\leq n$. If additionally, for each
$1\leq j\leq n$ and some $\sigma_{j}>0$,%
\begin{equation}
\lim_{k\rightarrow\infty}k^{-1}\sum_{l=1}^{k}\mathbb{V}\left[  v\left(
y_{lj}\right) |\mathbf{M} \right]  =\sigma_{j}\text{,}\label{18}%
\end{equation}
then $\sqrt{k}\left(  \hat{\delta}_{kj}-\bar{\delta}_{kj}\right)  $ converges
in distribution to a Normal random variable as $k \rightarrow \infty$.
\end{lemma}

Lemma \ref{Lm:AveVar} shows that $\hat{\mathbf{D}}_{k}$ in (\ref{23}) with $\hat{\delta}_{kj}$ defined by
\eqref{19} satisfies
$\lim\limits_{k\rightarrow\infty}\left\Vert\hat{\mathbf{D}}_{k}-\mathbf{D}_{k}\right\Vert \stackrel{\mbox{a.s.}}{=}0$.
Note that the first assertion in Lemma \ref{Lm:AveVar}, i.e.,
$\lim\limits_{k\rightarrow\infty}\left\vert \hat{\delta}_{kj}-\bar{\delta
}_{kj}\right\vert =0$ a.s.,
clearly applies to $y_{ij}$ that follow NEFs with QVF when their corresponding $\theta_{ij}$
are in a set $\Omega_0$ described in  Corollary \ref{Cor:GenToExp}.
We remark that requiring the closure of $\Omega_0$ to be in the
interior of $\Omega$ is not restrictive, since in practice the
$\theta_{ij}$'s are not the boundary points of $\Omega$.

\subsection{Simultaneously Consistently Estimating the Latent Linear Space and Its Dimension}
We are ready to present a consistent estimator of the latent linear space $\Pi_{\mathbf{M}}$:

\begin{enumerate}

\item Set $\mathbf{\hat{R}}_{k}=k^{-1}\mathbf{Y}^{T}\mathbf{Y}-\mathbf{\hat
{D}}_{k}$ as (\ref{23}) with $\hat{\delta}_{kj}$ as in (\ref{19}).

\item Estimate $r$ as $\hat{r}$ using \eqref{eq:rankest} as given in \autoref{Prop:EstRank}.

\item From the spectral decomposition $\mathbf{\hat{R}}_{k}=\mathbf{VKV}^{T}$
where $\mathbf{V}^{T}\mathbf{V}=\mathbf{I}$ and $\mathbf{K}=\diag\left\{
\hat{\alpha}_{k,i}\right\}  _{i=1}^{n}$, pick $\hat{r}$ columns $\left\{
\mathbf{\hat{u}}_{k,i}\right\}  _{i=1}^{\hat{r}}$ of $\mathbf{V}$
corresponding to the $\hat{r}$ largest $\left\{  \hat{\alpha}_{k,i}\right\}
_{i=1}^{\hat{r}}$.

\item Set
\begin{equation}
\hat{\Pi}_{\mathbf{M}}=\left\langle \left\{  \mathbf{\hat{u}}_{k,i}\right\}
_{i=1}^{\hat{r}}\right\rangle \label{65a}%
\end{equation}
to be the estimate of $\Pi_{\mathbf{M}}$.  Note that $\mathbf{\hat{M}}=\left(
\mathbf{\hat{u}}_{k,1},...,\mathbf{\hat{u}}_{k,\hat{r}}\right)  ^{T}$ can be
regarded as an estimator of $\mathbf{M}$ even though it is not our focus.
\end{enumerate}

The above procedure is supported by the following theorem, whose proof is straightforward and omitted. More specifically,
$\hat{\Pi}_{\mathbf{M}}$ in \eqref{65a} consistently estimates $\Pi_{\mathbf{M}}$.
\begin{theorem}
\label{Cor:AdjSampleCov}Under the assumptions of \autoref{Thm:AsympReducBzero}
and Lemma \ref{Lm:AveVar}, $\lim_{k\rightarrow\infty}\left\Vert
\mathbf{\hat{R}}_{k}-\mathbf{H}\right\Vert =0$ a.s.
and $\lim\limits_{k\rightarrow\infty} \left\langle \left\{  \mathbf{\hat{u}}_{k,i}\right\}
_{i=1}^{r}\right\rangle \triangle\Pi_{\mathbf{M}}=\varnothing $ a.s.. If
additionally the conditions of \autoref{Prop:EstRank} are satisfied, then equation
\eqref{eq:rankest} holds and
$\lim_{k\rightarrow\infty}\Pr\left( \left\{ \hat{\Pi}_{\mathbf{M}}\triangle\Pi_{\mathbf{M}} = \varnothing\right\} \right)=1$.
\end{theorem}

\subsection{Normal Distribution with Unknown Variances \label{sec:Leek}}
One of the exponential family distributions we considered above is $y_{ij} | \theta_{ij} \sim \mathsf{Normal}(\theta_{ij}, 1)$.  Suppose instead we assume that $y_{ij} | \theta_{ij} \sim \mathsf{Normal}(\theta_{ij}, \sigma^2_i)$, where $\sigma^2_1, \ldots, \sigma^2_k$ are unknown. \cite{Leek:2011} studies this important case, and obtains several results related to ours.   Let $a_1 \geq a_2 \geq \ldots \geq a_n$ be the $n$ ordered singular values resulting from the singular value decomposition (SVD) of $\mathbf{Y}$. If we regress the top $t-1$ right singular vectors in this SVD from each row of $\mathbf{Y}$, this yields total residual variation that is of proportion
$(\sum_{j=t}^n a_j^2)/(\sum_{j'=1}^n a_{j'}^2)$
to the total variation in $\mathbf{Y}$.  In order to estimate $\sigma^2_{\mbox{average}} = k^{-1} \sum_{j=1}^k \sigma^2_j$, \cite{Leek:2011} employs the estimate
$$
\hat{\sigma}^2_{\mbox{average}} = \frac{1}{k(n-t)} \frac{\sum_{j=t}^n a_j^2}{\sum_{j'=1}^n a_{j'}^2} \left\Vert \mathbf{Y} \right\Vert^2.
$$
Using our notation, \cite{Leek:2011} then sets $\hat{\mathbf{D}}_{k} = \hat{\sigma}^2_{\mbox{average}} \mathbf{I}$, where $\mathbf{I}$ is the $n \times n$ identity matrix, and proceeds to estimate $\Pi_{\mathbf{M}}$ based on $k^{-1} \mathbf{Y}^T \mathbf{Y} - \hat{\mathbf{D}}_{k}$ as we have done.  However, it must be the case that $t > r$ in order for $\hat{\sigma}^2_{\mbox{average}} $ to be well-behaved, so the assumptions and theory in \cite{Leek:2011} have several important differences from ours.  We refer the reader to \cite{Leek:2011} for specific details on this important case.  We note that taking our results together with those of \cite{Leek:2011}, this covers a large proportion of the models utilized in practice.

\section{Letting the Latent Variable Coefficients $\mathbf{\Phi}$ Be Random}  \label{sec:ranPhi}
We now discuss the case where $\mathbf{\Phi}$ is random but then conditioned.
Assume that $\boldsymbol{\Phi}$ is a random matrix with entries $\phi_{ij}$
defined on the probability space $\left(  \Omega_{1}%
,\mathcal{F}_{1},\mathbb{P}_{1}\right)  $, and that
the entries $y_{ij}$ of $\mathbf{Y}$, conditional on $\mathbf{\Phi}$ and
$\mathbf{M}$, are defined on the probability space $\left(  \Omega
,\mathcal{F},\mathbb{P}\right)  $.
Rather than model \eqref{7}, consider the following model:
\begin{equation}
\theta_{ij}=\mathbb{E}\left[  y_{ij}|\boldsymbol{\Phi},\mathbf{M}\right]  =\left(  \boldsymbol{\Phi
}\mathbf{M}\right)  \left(  i,j\right)  \text{ or }\boldsymbol{\Theta
}=\mathbb{E}\left[  \mathbf{Y|}\boldsymbol{\Phi},\mathbf{M}\right]
=\boldsymbol{\Phi}\mathbf{M}\text{.}\label{eqa4}%
\end{equation}
Suppose assumption A4) holds (see \eqref{11}) and:
\begin{description}
\item[A1$'$)] $1 \leq \rank\left(  \mathbf{M}\right)  =r<n$ and $n$ is finite; conditional on $\mathbf{M}$
and $\boldsymbol{\Phi}$, $\left\{  y_{ij}\right\}
_{i,j}$ are jointly independent with variance $\mathbb{V}\left[
y_{ij}|\mathbf{M},\boldsymbol{\Phi}\right]  =\delta_{ij}$ such that $\sup\limits_{k}\max\limits_{i,j}\mathbb{E}
\left[  y_{ij}^{8}|\mathbf{M},\boldsymbol{\Phi}\right]  \leq C$ (which implies $\sup\limits_{k}\max\limits
_{i,j}\mathbb{E}\left[  y_{ij}^{4}|\mathbf{M},\boldsymbol{\Phi}\right]  \leq C$), where $\mathbb{E}$ and $\mathbb{V}$ are the expectation and variance wrt $\mathbb{P}$.

\item[A2$'$)] Either A2) holds $\mathbb{P}_{1}$-a.s., i.e.,
$\mathbb{P}_{1}\left(\sup_{k \geq 1}\max_{1\leq i \leq k}\left\Vert \boldsymbol{\phi}_{i}\right\Vert \leq C\right)=1$ and
\begin{equation}
\mathbb{P}_{1}\left(  \left\Vert k^{-1}\boldsymbol{\Phi}^{T}\boldsymbol{\Phi
}-\mathbf{W}\right\Vert =c_{k},\exists c_{k}\rightarrow0,\exists
\mathbf{W}_{r\times r}>0\right)  =1,\label{eqa2}
\end{equation}
or A2) holds in probability $\mathbb{P}_{1}$, i.e., as $k \rightarrow \infty$,
$\mathbb{P}_{1}\left(\sup_{k \geq 1}\max_{1\leq i \leq k}\left\Vert \boldsymbol{\phi}_{i}\right\Vert \leq C \right) \rightarrow 1$ and
\begin{equation}
\mathbb{P}_{1}\left(  \left\Vert k^{-1}\boldsymbol{\Phi}^{T}\boldsymbol{\Phi
}-\mathbf{W}\right\Vert =c_{k},\exists c_{k}\rightarrow0,\exists
\mathbf{W}_{r\times r}>0\right)  \rightarrow1\text{.}\label{eqa3}%
\end{equation}

\item [A3$'$)] Conditional on $\mathbf{M}$, $k^{-1}\sum\limits_{i=1}^{k}\mathbb{E}\left[  e_{ij}%
^{4}|\mathbf{M}\right]  $, $k^{-1}\sum\limits_{i=1}^{k}\boldsymbol{\phi}_{i}\mathbb{E}\left[
e_{ij}^{3}|\mathbf{M}\right]  $ and $k^{-1} \sum_{i=1}^{k} \boldsymbol{\phi}_i \mathbf{m}^{j} \boldsymbol{\phi}_i \mathbf{m}^l\boldsymbol{\Delta}_i $ for any $1 \leq j \leq l \leq n$
are all convergent $\mathbb{P}_{1}$-a.s. as $k \rightarrow \infty$.
\end{description}

Note that assumptions A1$'$), A2$'$) and A3$'$) are the probabilistic versions of assumptions A1), A2 and A3) that also account for the randomness of $\mathbf{\Phi}$. Recall assumption A2) when $\boldsymbol{\Phi}$ is
deterministic, i.e., for some $\mathbf{W}_{r\times r}>0$ and some non-negative
sequence $c_{k}\rightarrow0$, $\left\Vert k^{-1}\boldsymbol{\Phi}^{T}\boldsymbol{\Phi}-\mathbf{W}\right\Vert
=c_{k}$. Assumption A2) implies that (see also Lemma \ref{Lm:UnifConv}):
\begin{equation}
\sup_{k\geq1}\max_{1\leq i\leq k}\left\Vert \boldsymbol{\phi}_{i}\right\Vert
\leq C\text{ \ \ and \ }\sup_{k\geq1}\max_{1\leq i\leq k,1\leq j\leq
n}\left\vert \theta_{ij}\right\vert \leq C. \label{eqa1}%
\end{equation}
These two uniform boundedness results in equation (\ref{eqa1}) are then used to show
the a.s.\ convergence in Theorem \ref{Thm:AsympReducBzero} which induces the a.s.\ convergence in
Corollary \ref{Lm:ConvEigensys}, the validity of Lindeberg's condition as \eqref{81a} that induces Proposition \ref{Prop:CLTRandomPart},
convergence in probability in equations \eqref{26} and \eqref{20} that induces Theorem \ref{Prop:EstRank}, Corollary \ref{Cor:GenToExp}, and Theorem \ref{Cor:AdjSampleCov}.

Let $\tilde{N}_{1}\in\mathcal{F}_{1}$ be such that (\ref{eqa2}) does not hold,
then $\mathbb{P}_{1}\left(  \tilde{N}_{1}\right)  =0$. On the other hand, if
(\ref{eqa3}) holds, then for any positive constants $C$ and $\tilde{\varepsilon}$ there exists
a $k_0=k_0\left(C,\tilde{\varepsilon}\right)$ such that the set
\[
N_{1}^{\ast}\left(C,\tilde{\varepsilon}\right)=\left\{  \omega\in\Omega_{1}:\sup\nolimits_{k \geq 1}\max\nolimits_{1\leq i\leq k
}\left\Vert \boldsymbol{\phi}_{i}\right\Vert > C\right\}
\]
satisfies
\[
\mathbb{P}_{1}\left( N_{1}^{\ast}\left(C,\tilde{\varepsilon}\right)\right)
 <\tilde{\varepsilon}
 \]
whenever $k > k_0$.
Now if (\ref{eqa2}) holds, then the results on the a.s. convergence and on
convergence in probability for $y_{ij}|\mathbf{M}$ remain true with respect to $\mathbb{P}$
when $y_{ij}|\mathbf{M}$ is replaced by $y_{ij}|\left(\boldsymbol{\Phi},\mathbf{M}\right)  $, as long as
each $\phi_{ij}\left(  \omega\right)$ is such that $\omega\in\Omega
_{1}\setminus\tilde{N}_{1}$.
In contrast, if (\ref{eqa3}) holds, then the
results on the a.s.\ convergence for $y_{ij}|\mathbf{M}$
reduce to convergence in probability in $\mathbb{P}$. But those on convergence in probability
for $y_{ij}|\mathbf{M}$ remain true with respect to $\mathbb{P}$
when $y_{ij}|\mathbf{M}$ is replaced by $y_{ij}|\left(  \boldsymbol{\Phi},\mathbf{M}\right)  $,
as long as each $\phi_{ij}\left(  \omega\right)  $ is with
$\omega\in\Omega_{1}\setminus N_{1}^{\ast}\left(C,\tilde{\varepsilon}\right)$
(where $\tilde{\varepsilon}$ can be chosen to be small).
Therefore, the results (except Lemmas \ref{Lm:NEFQVF} and \ref{Lm:AveVar}) obtained in the previous sections
hold with their corresponding probability statements, whenever,
for some $\mathbf{W}_{r\times r}>0$,
\begin{equation}\label{eqa5}
\left\Vert k^{-1}\boldsymbol{\Phi}^{T}\boldsymbol{\Phi
}-\mathbf{W}\right\Vert = c_k \rightarrow 0
\end{equation}
holds a.s.\ or in probability in $\mathbb{P}_{1}$, and they
allow $\boldsymbol{\Phi}$ to be random (and conditioned) or deterministic.
Statistically speaking, an event with very small or zero $\mathbb{P}_{1}$ probability is very unlikely to
occur. This means that the practical utility of these results
is not affected when \eqref{eqa5} holds in the sense of \eqref{eqa2} or \eqref{eqa3}
when $k$ is large.

Finally, we remark that Lemmas \ref{Lm:NEFQVF} and \ref{Lm:AveVar} are independent of the assumptions A1), A2) and A3), and
of A1$'$), A2$'$) and A3$'$). In other words, for model \eqref{eqa4}, Lemma \ref{Lm:NEFQVF} remains valid,
and so does Lemma \ref{Lm:AveVar} when the assumptions involved there are replaced by those on
$y_{ij}|\left(  \boldsymbol{\Phi},\mathbf{M}\right)$.

\section{A Simulation Study \label{Sec:SimThyEstM}}
We conducted a simulation study to demonstrate the validity of our theoretical results.
The quality of $\hat{\Pi}_{\mathbf{M}}$ is measured by%
\[
d\left(  \mathbf{M},\mathbf{\hat{M}}\right)  =\left(  \sqrt{n\hat{r}}\right)
^{-1}\sqrt{\left\Vert \mathbf{M}^{T}-\mathbf{\hat{M}}^{T}\mathbf{M}%
_{\mathbf{V}}\right\Vert ^{2}+\left\Vert \hat{\mathbf{M}}^{T}-\mathbf{M}%
^{T}\mathbf{V}_{\mathbf{M}}\right\Vert ^{2}},
\]
where $\hat{r}$ is an estimate of $r$, $\mathbf{M}_{\mathbf{V}}=\mathbf{\hat
{M}M}^{T}$ and $\mathbf{V}_{\mathbf{M}}=\left(  \mathbf{MM}^{T}\right)
^{-1}\mathbf{M\hat{M}}^{T}$. In fact, $d\left(  \mathbf{M},\mathbf{\hat{M}}\right)$ measures the
difference between the orthogonal projections (as linear operators) induced by
the rows of $\mathbf{\hat{M}}$ and those of $\mathbf{M}$ respectively, and $d\left(  \mathbf{M}%
,\mathbf{\hat{M}}\right)  =0$ if and only if $\hat{\Pi}_{\mathbf{M}}%
=\Pi_{\mathbf{M}}$. We propose a
rank estimator $\hat{r}$, directly applied to $\mathbf{\hat{R}}_{k}$, based on
\autoref{Prop:EstRank}. Specifically, $\tilde{\tau}_{k}=\tilde{c}_{k}k^{-\eta}$
for some positive $\tilde{c}_{k}$ and $\eta$ is used to
scale the eigenvalues of $\mathbf{\hat{R}}_{k}$, for which
$\tilde{c}_{k}$ (dependent on $k$) is determined by a strategy of
\cite{Hallin:2007} (that was also used by \citealp{Leek:2011}); for more details,
we refer the reader to these two references. We do so since the assertions in \autoref{Prop:EstRank}
are valid only for $k=\infty$, assumption A3) may not be satisfied in our simulation study, and
the unknown constants in equation (\ref{26}) for the speed of convergence need to be estimated. However, we caution
that, if the scaling sequence $\tilde{\tau}_{k}$ in \autoref{Prop:EstRank} defined via
\eqref{25} is not well estimated to capture the speed of convergence of the eigenvalues of $\hat{\mathbf{R}}_k$,
$\hat{r}$ defined by \eqref{eq:rEstDef} as an estimate of $r$ can be inaccurate; see our simulation study results.

\subsection{Simulation Study Design}

The settings for our simulation study are described below:

\begin{enumerate}
\item We consider $n=15,100,200$, $k= 10^{3},5\times10^{3},10^{4},10^{5}$ and
$r=1,2,3,4,5,6,8,10,12$.

\item The mean matrix $\boldsymbol{\Theta}=\boldsymbol{\Phi}\mathbf{M}$ and the observed data $y_{ij} | \theta_{ij}$ are generated
within the following scenarios:

\begin{enumerate}
\item $y_{ij} | \theta_{ij} \sim \mbox{Normal}(\theta_{ij},1)$, $\boldsymbol{\Phi}\left(i,j\right)  =\phi_{ij} \iid \mathsf{Normal}\left(  0,1\right)$, and $\mathbf{M}\left(  i,j\right)=m_{ij} \iid \mathsf{Uniform}\left(1,10\right)$.

\item $y_{ij} | \theta_{ij} \sim \mathsf{Poisson}\left(  \theta_{ij}\right)$,
$\phi_{ij} \iid \mathsf{Chi}$-$\mathsf{square}\left(  9,1\right)$
with degrees of freedom $9$ and non-centrality $1$, and $m_{ij} \iid \mathsf{Uniform}\left(  1,5\right)$.

\item $y_{ij} | \theta_{ij} \sim \mathsf{Binomial}\left(s,\theta_{ij}\right)$ with $s=20$ and $\phi_{ij} \iid \mathsf{Uniform}\left(
0.05,0.95\right)$. $\mathbf{M}$ is such that $m_{ij} = 1$ for $1 \leq i=j \leq r$, $m_{ij} = r^{-1}$ for $1\leq i\leq
r$ and $r+1\leq j\leq n$, and $m_{ij}=0$ otherwise.

\item $y_{ij} | \theta_{ij} \sim \mathsf{NegBin}\left(s,\theta_{ij}\right)$ with $s=10$, $\phi_{ij} \iid \mathsf{Uniform}\left(0.5,2\right)$, and $m_{ij} \iid \mathsf{Uniform}\left(  0.3,1.5\right)$.

\item $y_{ij} | \theta_{ij} \sim \mathsf{Gamma}\left(s,\theta_{ij}\right)$ with shape $s=10$ and mean $\theta_{ij}$, $\phi_{ij} \iid \mathsf{Uniform}\left(  0.5,2\right)$, and $m_{ij} \iid \mathsf{Uniform}\left(  0.3,1.5\right)$.
\end{enumerate}

\item For each combination of values in Step 1 and distributions in Step 2 above, the data matrix $\mathbf{Y}$ is generated
independently $100$ times and then the relevant statistics are calculated.
\end{enumerate}

For each simulated data set, we measure the quality of $\mathbf{\hat{D}}_{k}$ by $\rho_{k}
 =\max\limits_{1\leq j\leq n}\left\vert
\hat{\delta}_{kj}  -\bar{\delta}_{kj}
\right\vert $, that of $\hat{\Pi}_{\mathbf{M}}$ by  $d\left(  \mathbf{M},\mathbf{\hat{M}}\right)$,
and we record $\hat{r}$.

\begin{table}[tbp]
\centering
\caption[]{Assessment of the estimator $\hat{r}$.  Data were simulated under model \eqref{7} with $n=15$ and $r=5$ under the five distributions listed.  Shown is the number of times that $\hat{r}=r$ among $100$ simulated data sets for each scenario.  Also shown in parentheses is the number of times that $\hat{r} < r$, if any instances occurred. Results from several additional scenarios are considered in \autoref{AppC:ResSimu}.}
\label{TbA1}
\begin{center}
\begin{tabular}{|l|l|l|l|l|l|l|}
  \hline
     $k$      & Binomial & Gamma & NegBin & Normal & Poisson \\
  \hline
       1000  &            3 (97) &    24 (4) &       3 &        100 &      84 (6) \\
   5000  &   95 (5) &    30 (9) &        4 &          96 &      94 \\
       10,000 &              100 &        33 (22)&        27 (9) &     96 &      89 \\
       100,000 &              100 &        90 &            43 &        99 &      94 \\
   \hline
\end{tabular}
\end{center}
\end{table}

\subsection{Simulation Study Results}
\autoref{FigA1} and \autoref{TbA1} display the performance of the
nonparametric estimator $\hat{\Pi}_{\mathbf{M}}$ of $\Pi_{\mathbf{M}}$, that of $\mathbf{\hat{D}}_{k}$, and
that of $\hat{r}$ when $n=15$ and $r=5$; for other values of $n$ and $r$, the performance of $\hat{r}$
is provided in \autoref{AppC:ResSimu} and that of $\hat{\Pi}_{\mathbf{M}}$ in \autoref{AppD:ResSimu}. The following
conclusions can be drawn from the simulation study:

\begin{figure}[tpbh]
\centering
\includegraphics[height=0.56\textheight]{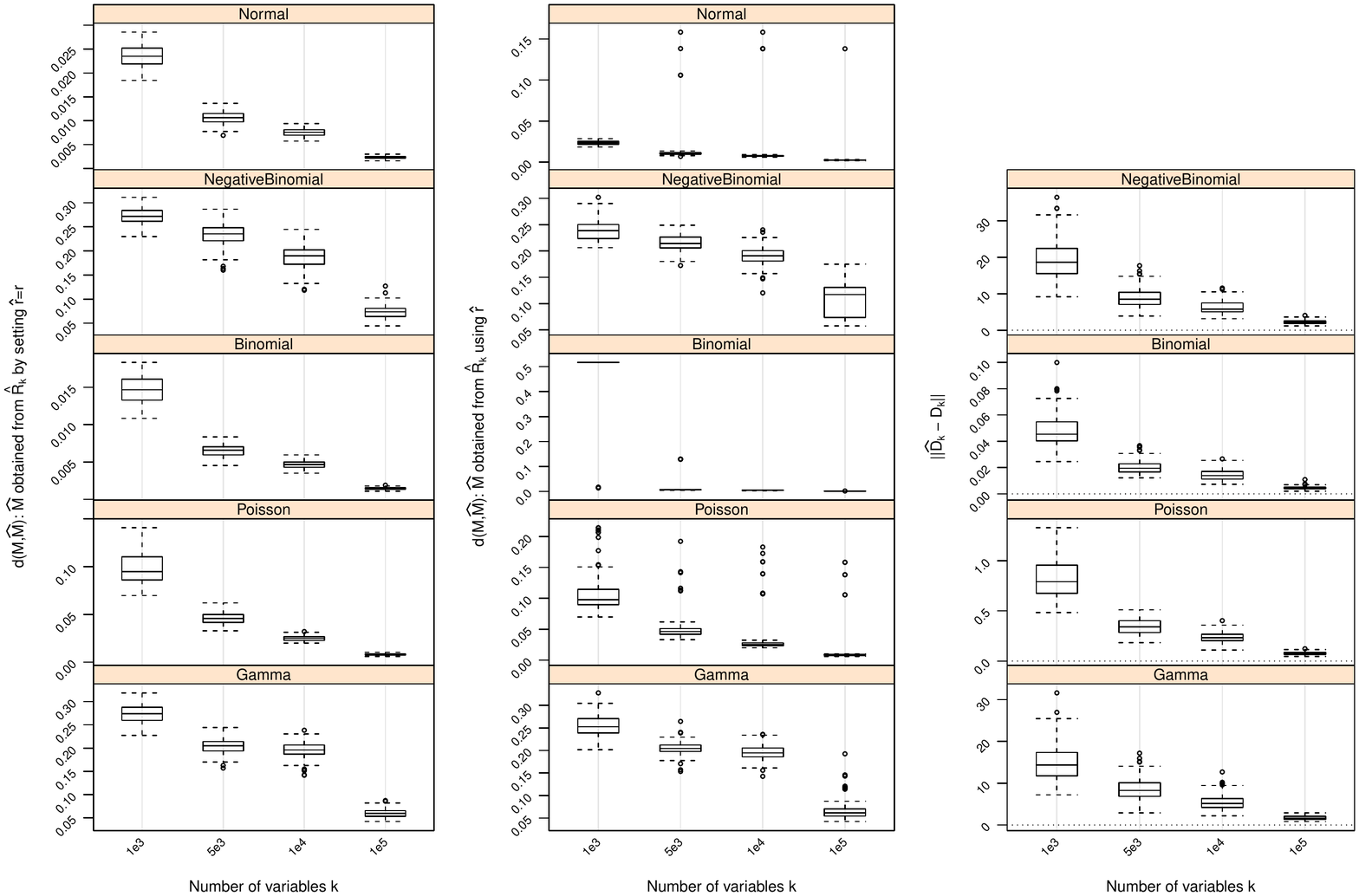}
\caption[Nonparametric Estimation of $\Pi_{ \mathbf{M}} $: $r=5$]{
Performance of the nonparametric estimator $\hat{\Pi}_{ \mathbf{M}} $ of $\Pi_{ \mathbf{M}} $ when $n=15$ and $r=5$. Column 1: Boxplots of the difference between the row spaces spanned by $\mathbf{M}$ and $\mathbf{\hat{M}}$ as measured by $d(\mathbf{M},\mathbf{\hat{M}})$ when the true dimension of $\mathbf{M}$ is utilized to form $\mathbf{\hat{M}}$ (i.e., setting $\hat{r}=r$). Column 2: Boxplots of $d(\mathbf{M},\mathbf{\hat{M}})$ when using the proposed estimator $\hat{r}$ of the row space dimension in forming $\mathbf{\hat{M}}$.  Column 3:  An assessment of the estimate $\mathbf{\hat{D}}_k$ of $\mathbf{D}_k$, where the latter term is the average of the column-wise variances of $\mathbf{Y}$.  The difference is measured by $\Vert \mathbf{\hat{D}}_k - \mathbf{D}_k \Vert = \max\limits_{1\leq j\leq n} \vert \hat{\delta}_{kj}-\bar{\delta}_{kj} \vert $.  Results from several additional scenarios are shown in \autoref{AppD:ResSimu}.
\label{FigA1}}
\end{figure}

\begin{enumerate}
\item $\hat{\Pi}_{\mathbf{M}}$ with $\hat{r}$ set as the true $r$ approximates $\Pi_{\mathbf{M}%
}$ with increasing accuracy as measured by the difference between their induced
 orthogonal projections $d\left(  \mathbf{M},\mathbf{\hat{M}}\right)$ when $k$ gets larger.
 In all settings, strong trends of convergence of $\hat{\Pi}_{\mathbf{M}}$ to $\Pi_{\mathbf{M}}$
 can be observed, even when $\hat{r}$ is used. However, the speed of convergence can
 be slightly different for different settings
  due to the hidden constants in \eqref{2} and \eqref{26}.

\item As $k$ gets larger $\mathbf{\hat{D}}_{k}$ becomes more accurate, and strong
trends of convergence of $\mathbf{\hat{D}}_{k}$ to $\mathbf{D}_{k}$ can be observed in
all settings. This is similar to the behavior of $\hat{\Pi}_{\mathbf{M}}$.
However, the accuracy of $\mathbf{\hat{D}}_{k}$ in estimating $\mathbf{D}_{k}$
does not seem to have a drastic impact on $\hat{\Pi}_{\mathbf{M}}$, since $\mathbf{\hat{D}}_{k}$ induces
a shift by a diagonal matrix to the matrix $k^{-1}\mathbf{Y}^{T}\mathbf{Y}$ and such
a shift does not necessarily have a huge impact on the leading eigenvectors of $\mathbf{\hat{R}}_{k}$ and
hence on $\hat{\Pi}_{\mathbf{M}}$. We do not report the performance of $\mathbf{\hat{D}}_{k}$ in the scenarios where
$y_{ij} | \theta_{ij} \sim \mathsf{Normal}(\theta_{ij},1)$ since in this case
$\delta_{ij}=1$ and $\mathbf{D}_{k}=\mathbf{I}$, and $\mathbf{\hat{D}}_{k}=\mathbf{I}$ has been set.

\item In all settings, $\hat{r}$ can under-
or over-estimate $r$, and as $k$ increases $\hat{r}$ becomes more accurate.
However, when $\hat{r} > r$ it does not reduce the
accuracy of the estimate $\hat{\Pi}_{\mathbf{M}}$ when $\hat{r}$ is used to pick the number
of leading eigenvectors of $\mathbf{\hat{R}}_{k}$ to estimate $\Pi_{\mathbf{M}}$. In fact, when $\hat{r} >r$, additional linearly independent eigenvectors $\hat{\mathbf{u}}_{k,i}$ for $i=1,\ldots,r$ may be used to span $\Pi_{\mathbf{M}}$,
giving better estimate than using the true $r$. When $\hat{r} < r$, the accuracy of $\hat{\Pi}_{\mathbf{M}}$ is reduced, since in this case the original row space $\Pi_{\mathbf{M}}$ can not be sufficiently spanned by $\hat{\Pi}_{\mathbf{M}}$. This is clearly seen from the plots
on performance of $\hat{\Pi}_{\mathbf{M}}$.

\item The scaling sequence $\left\{  \tilde{\tau}%
_{k}\right\}  _{k\geq1}$ plays a critical role on the accuracy of
$\hat{r}$ as an estimator of $r$, since it decides where to
``cut'' the spectrum of $\mathbf{\hat{R}}_{k}$ for finite $k$ to give $\hat{r}$, where
numerically all eigenvalues of $\mathbf{\hat{R}}_{k}$ are non-zero. In all settings,
the non-adaptive choice of the sequence $\left\{  \tilde{\tau}%
_{k}\right\}  _{k\geq1}$ with $\tilde{\tau}_{k}=\tilde{c}_{k}k^{-1/3}$
has been used, and it can cause inaccurate scaling of the
spectrum of $\hat{\mathbf{R}}_{k}$ and hence inaccurate estimate of $r$.
This explains why $r$ has been under-estimated when $y_{ij}$'s follow
Binomial distributions, $n\in\left\{100,200\right\}$ and $r\geq 2$, since in these cases
the magnitudes of the eigenvalues of $\mathbf{\hat{R}}_{k}$ have more complicated behavior,
and $\tilde{\tau}_{k}$ are too large compared the magnitudes of these eigenvalues;
see Tables \ref{tab:a1}--\ref{tab:a3} in \autoref{AppC:ResSimu} for results when $n=100$ and those when $n=200$.
We found that when $y_{ij}$ follow Binomial distributions in the simulation
study, in order to accurately estimate $r$,
$\tilde{\tau}_{k}=O\left(k^{-1/1.1}\right)$ should be set for $n=100$ and that
$\tilde{\tau}_{k}=O\left(k^{-1/1.5}\right)$ should be set for $n=200$.
The non-adaptive choice of $\tilde{\tau}_{k}$ also explains why
$\hat{r}$ over-estimates $r$ when $y_{ij}$'s follow the Negative Binomial distributions
or gamma distributions in the simulation study, since these two types of
distributions are more likely to generate
outliers that counterbalance the speed of concentration as $k$ gets larger, and
affect the separation of the spectrum of the limiting matrix $\hat{\mathbf{R}}_{k}$. In other words, for these two cases,
$\tilde{\tau}_{k}$ are too small in magnitudes in order to scale up the spectrum of
$\hat{\mathbf{R}}_{k}$ to the point of separation in order to estimate $r$. In general,
the magnitude of $n$ plays a role in the asymptotic property of $\hat{\mathbf{R}}_{k}$
as $k\rightarrow \infty$, and it affects the speed of convergence of $\hat{\mathbf{R}}_{k}$
through the hidden constants in \eqref{26} even when all needed assumptions are satisfied for \autoref{Prop:EstRank}.
This explains why larger $n$ does not necessarily
induce more accurate $\hat{r}$ when $r$ increases but $k$ does not increase at a compatible speed when
$k$ is finite; see Tables \ref{tab:a1}--\ref{tab:a3} in \autoref{AppC:ResSimu}.
However, deciding the hidden constants in \eqref{26} is usually very hard,
and accurately estimating $r=\rank\left(  \mathbf{M}\right)  $, being also the number of
factors in factor models, is in general a very challenging problem.
\end{enumerate}

\begin{figure}[p]
\centering
\includegraphics[width=\textwidth]{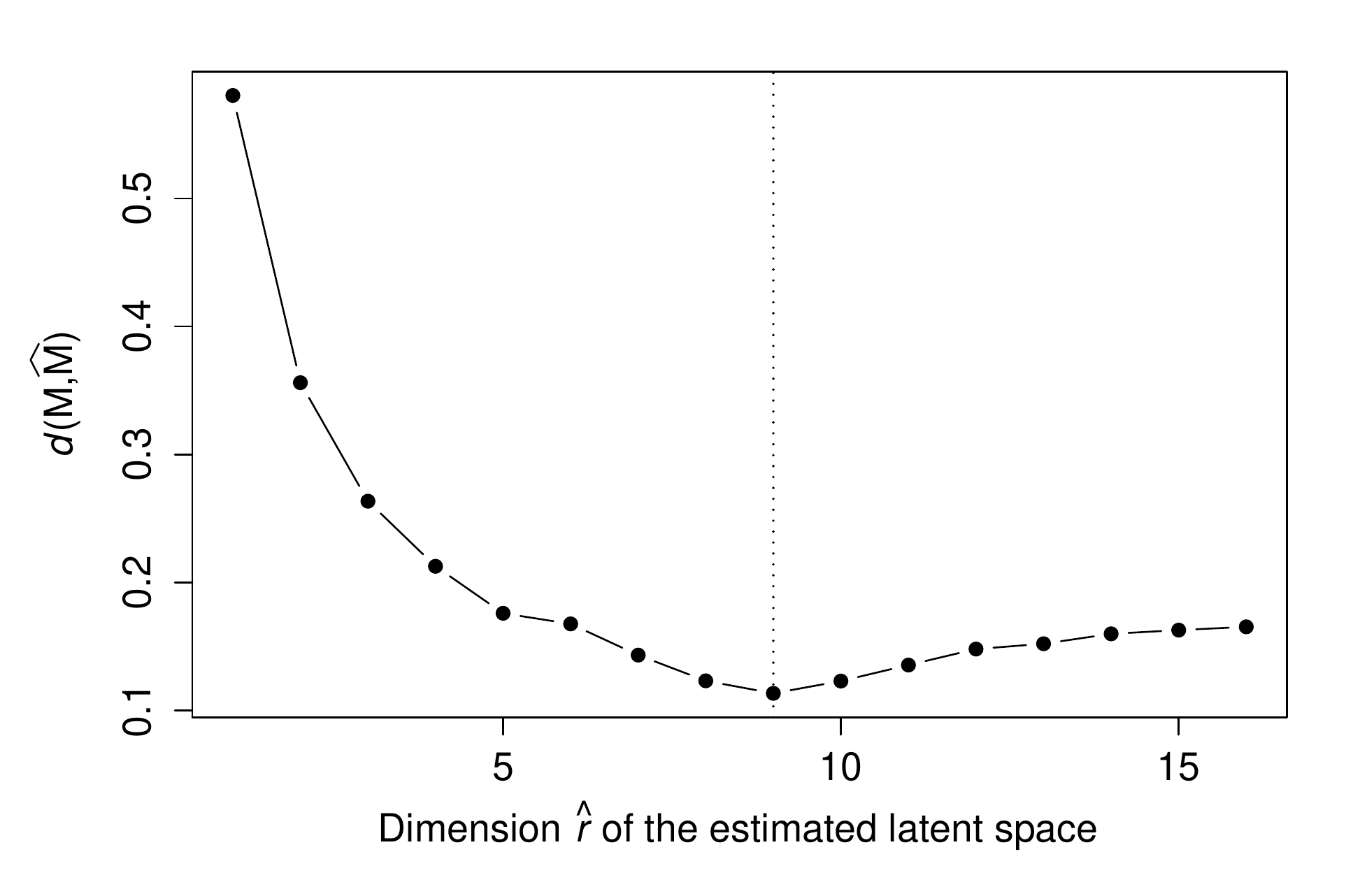}
\caption[Example: distance as r changes]{
A measure of how well $\hat{\mathbf{M}}$ captures the row space of $\mathbf{M}$ as the dimension $\hat{r}$ of $\hat{\mathbf{M}}$ increases in the RNA-seq data set. This measure $d\left( \mathbf{M}, \mathbf{\hat{M}}\right)$ is plotted versus $\hat{r}$ over $\hat{r} \in \{1, 2, \ldots, 16\}$.  The true dimension of the row space of $\mathbf{M}$ is $r=9$, shown by vertical dotted line.}
\label{FigEgA}
\end{figure}

\begin{figure}[p]
\centering
\includegraphics[width=\textwidth]{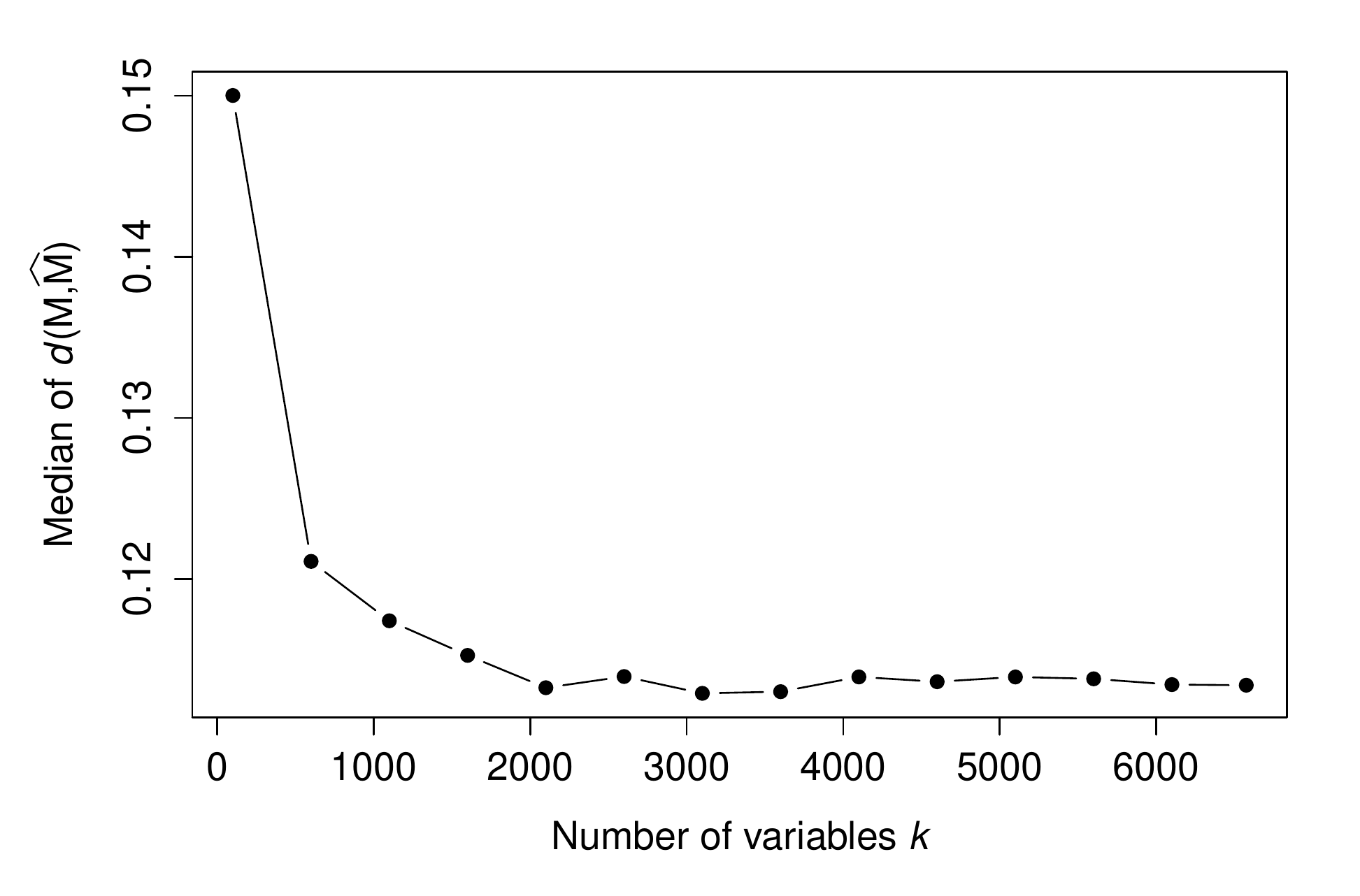}
\caption[Example: distance as k changes]{
A measure of how well $\hat{\mathbf{M}}$ captures the row space of $\mathbf{M}$ as the number of variables $k$ increases in the RNA-seq data set.  For each value of $k$, we randomly sampled $k$ rows from the RNA-seq data matrix and computed $d\left( \mathbf{M}, \mathbf{\hat{M}}\right)$ as defined in the text; this was repeated 50 times and the median was plotted.}
\label{FigEgB}
\end{figure}

\section{Application to an RNA-Seq Study}
\label{sec:rnaseq}

In \citet{Robinson2015}, we measured genome-wide gene expression in the budding yeast {\em Saccharomyces cerevisiae} in a nested factorial experimental design that allowed us to carefully partition gene expression variation due to both biology and technology.  The technology utilized to measure gene expression is called ``RNA-seq'', short for RNA sequencing.  This technology provides a digital measure of gene expression in that mRNA molecules are discretely sequenced and therefore counted \citep{Wang:2009}.  The resulting $6575 \times 16$ matrix of RNA-seq counts represents gene expression measurements on 6575 genes across 16 samples.  We also have a design matrix of dimension $9 \times 16$ that captures the experimental design and apportionment of variation throughout the data.   The data here are counts, and RNA-seq count data are typically modeled as overdispersed Poisson data \citep{McCarthy:2012fv}.  It can be verified on these data that, because of the experimental design, there is very little over-dispersion once the experimental design is taken into account; we therefore utilize the Poisson distribution in this analysis.

We set $\mathbf{Y}$ to be the $6575 \times 16$ matrix of RNA-seq counts and $\mathbf{M}$ the $9 \times 16$ design matrix, where each row of $\mathbf{M}$ is normalized to have unit Euclidean norm.  Ignoring our knowledge of $\mathbf{M}$, we applied the proposed estimator using the Poisson distribution formula to yield an estimate $\mathbf{\hat{M}}$ over a range of values of $\hat{r} \in \{1, 2, \ldots, 16\}$. In evaluating $\mathbf{\hat{M}}$, we utilized the measure $d\left( \mathbf{M}, \mathbf{\hat{M}}\right)$ defined  in \autoref{Sec:SimThyEstM}.

\autoref{FigEgA} shows how $d\left( \mathbf{M}, \mathbf{\hat{M}}\right)$ changes with the dimension $\hat{r}$ of $\mathbf{\hat{M}}$. It can be seen that the minimum value is at $\hat{r} = 9$ (where $r=9$ is the true dimension of $\mathbf{M}$).  We then randomly sampled rows of $\mathbf{Y}$ to quantify how $d\left( \mathbf{M}, \mathbf{\hat{M}}\right)$ changes as the number of variables $k$ grows.  \autoref{FigEgB} shows the median $d\left( \mathbf{M}, \mathbf{\hat{M}}\right)$ values over 50 random samplings for each $k$ value, where a convergence with respect to $k$ can be observed.  In summary, this analysis shows that the proposed methodology is capable of accurately capturing the linear latent variable space on real data that is extremely heteroskedastic and does not follow the Normal distribution.

\section{Discussion\label{Sec:Disc}}
We have proposed a general method to consistently estimate the low-dimensional, linear latent variable structure in a set of high-dimensional rv's. Further, by exploiting the intrinsic relationship between the moments of natural exponential family (NEF) distributions that have quadratic variance functions (QVFs), we are able to explicitly recover this latent structure by using just the second moments of the rv's even when these rv's have heteroskedastic variances. Empirical evidence confirms our theoretical findings and the utility of our methodology. Once the latent structure has been well estimated, the variable-specific coefficients of the latent variables can be estimated via appropriate estimation methods.

We point out that, under the same assumptions A1), A2), A3) and A4), the theoretical results in Sections \ref{Sec:NPEstM}, \ref{sec:estr} and \ref{Sec:AppToExpFamily} hold  for the unconditional model
\begin{equation}
\boldsymbol{\Theta
}=\mathbb{E}\left[  \mathbf{Y}\right]  =\boldsymbol{\Phi}\mathbf{M}\label{7a}
\end{equation}
when both $\boldsymbol{\Phi}$ and $\mathbf{M}$ are deterministic with $\rank\left(\mathbf{M}\right)=r$. In this case the conditional distribution of each $y_{ij}$ given $\mathbf{M}$
in model (\ref{7}), i.e.,
$\boldsymbol{\Theta
}=\mathbb{E}\left[  \mathbf{Y}|\mathbf{M}\right]  =\boldsymbol{\Phi}\mathbf{M}$
when $\mathbf{M}$ is a realization of a random matrix $\tilde{\mathbf{M}}$, becomes
the marginal distribution of $y_{ij}$ in model (\ref{7a}) and
the proofs proceed almost verbatim as those for model (\ref{7}).
Further, for the model
\begin{equation}
\boldsymbol{\Theta
}=\mathbb{E}\left[  \mathbf{Y}|\boldsymbol{\Phi}\right]  =\boldsymbol{\Phi}\mathbf{M}\label{7b}
\end{equation}
when $\boldsymbol{\Phi}$ is random but $\mathbf{M}$ is deterministic with $\rank\left(\mathbf{M}\right)=r$,
under the assumptions A1$'$), A2$'$), A3$'$) and A4),
similar arguments as those given in \autoref{sec:ranPhi} imply that the theoretical results obtained
in Sections \ref{Sec:NPEstM}, \ref{sec:estr} and \ref{Sec:AppToExpFamily} hold
when the probabilistic statements in the conclusions there are adjusted in the way detailed in \autoref{sec:ranPhi}.

We have observed that for certain NEF distribution configurations, estimating the rank of the linear space generated by the latent structure
can can be challenging, even with the theory we provided, if the scaling sequence to separate the limiting spectrum of the adjusted gram matrix of the data is not adaptively specified. The need for choosing an adaptive scaling sequence comes from the hidden constants that describe the asymptotic speed of convergence, and it can be a delicate task to do so in non-asymptotic settings. This is reflective of the more general challenge of estimating the dimension of a latent variable model.

Finally, we briefly point out that, when the $y_{ij}$ have pdf
$f(y_{ij};\theta_{ij},\psi)$ such that
$$
f(y;\theta,\psi)=h\left(  y,\psi\right)  \exp\left\{\psi^{-1}\left[\eta(\theta)y-g\left(  \eta\left(  \theta\right)  \right) \right] \right\}
$$
form an exponential dispersion family (see, e.g., \citealp{Jorgensen:1987}), an explicit estimator of $\mathbf{D}_{k}$ (and hence that of $\Pi_{\mathbf{M}}$) will require results beyond those provided in this work, even if the unknown dispersion parameter $\psi$ is constant.  We leave this case to future research.

\section*{Acknowledgements}
This research is funded by the Office of Naval Research grant N00014-12-1-0764 and NIH grant HG002913.

\bibliography{Intragression}
\bibliographystyle{natbib}

\clearpage
\appendix{}

\section{Relationship to Anandkumar {\em et al.} \label{sec:Anandkumar}}
\cite{Anandkumar:2012,Anandkumar:2015} consider a different class of probabilistic models than we consider. We establish this by rewriting their model in our notation. We consider the model
$$
\mathbb{E}\left[  \mathbf{Y}|\mathbf{M}\right]  =\boldsymbol{\Phi}_{k \times r} \mathbf{M}_{r \times n},
$$
where $\mathbf{Y}_{k \times n}$ is a matrix of $n$ observations on $k$ variables.  The data points $y_{ij}$ can take on a wide range of classes of variables, from Binomial outcomes, to count data, to continuous data (see \autoref{tab:vfunc}, for example).  Variable $i$ can be written as $\mathbf{y}_i = (y_{i1}, \ldots, y_{in})$, which is a $1 \times n$ vector.  In terms of this variable, our model assumes that
$$
\mathbb{E}\left[  \mathbf{y}_i |\mathbf{M}\right]  =\boldsymbol{\phi}_i \mathbf{M},
$$
where $\boldsymbol{\phi}_i \in \real^{1 \times r}$ is row $i$ of $\boldsymbol{\Phi}$.

\cite{Anandkumar:2012,Anandkumar:2015}, on the other hand, assume that $y_{ij} \in \{0,1\}^n$ with the restriction that $\sum_{j=1}^n y_{ij} = 1$.  This construction is meant to represent an observed document of text where there are $n$ words in the dictionary.  They consider a single document with an infinite number of words.  The vector $\mathbf{y}_i = (y_{i1}, \ldots, y_{in})$ tells us which of the $n$ words is present at location $i$ in the document.  When they let $k \rightarrow \infty$, this means the number of words in the document grows to infinity.

The model studied in \cite{Anandkumar:2012,Anandkumar:2015} is
$$
\mathbb{E}\left[  \mathbf{y}_i^T |\mathbf{m}\right]  = \mathbf{O}_{n \times r} \mathbf{m}_{r \times 1},
$$
where there are $r$ topics under consideration and the $r$-vector $\mathbf{m}$ gives the mixture of topics in this particular document.  Each column of $\mathbf{O}$ gives the multinomial probabilities of the $n$ words for the corresponding topic.  Note that the linear latent variable model $\mathbf{O}\mathbf{m}$ does not vary with $i$, whereas in our model it does.  Also, the dimensionality of the latent variable model is different than ours.  Therefore, this is a different model than we consider.

\cite{Anandkumar:2012,Anandkumar:2015} take the approach of calculating, projecting and decomposing the expectations of tensor products involving $(\mathbf{y}_1, \mathbf{y}_2)$ and $(\mathbf{y}_1, \mathbf{y}_2, \mathbf{y}_3)$, and then suggesting that these expectations can be almost surely estimated as $k \rightarrow \infty$ by utilizing the analogous sample moments.  In order to exactly recover $\mathbf{O}$ modulo a permutation of its columns, additional assumptions are made by \cite{Anandkumar:2012,Anandkumar:2015}, such as knowledge of the sum of the unknown exponents in the Dirichlet distribution in the LDA model of \cite{Blei:2003}.  It is also assumed that the number of topics $r$ is known.

\section{Proofs \label{AppB:proof}}

Since we are considering the conditional model \eqref{7}, i.e.,
$ \boldsymbol{\Theta
}=\mathbb{E}\left[  \mathbf{Y}|\mathbf{M}\right]  =\boldsymbol{\Phi}\mathbf{M} $, to maintain
concise notations we introduce the conditional expectation and variance operators
respectively as $\tilde{\mathbb{E}}\left[\cdot\right]=\mathbb{E}\left[\cdot|\mathbf{M}\right]$ and
$\tilde{\mathbb{V}}\left[\cdot\right]=\mathbb{V}\left[\cdot|\mathbf{M}\right]$. In the proofs,
unless otherwise noted, the random vectors are conditioned on $\mathbf{M}$ and
the arguments for these random vectors are conditional on $\mathbf{M}$.

\subsection{Proof of \autoref{Thm:AsympReducBzero}}

Denote
the terms in the order in which they appear in the expansion%
\begin{align*}
k^{-1}\mathbf{Y}^{T}\mathbf{Y}  & =k^{-1}\boldsymbol{\Theta}^{T}\boldsymbol{\Theta}%
+k^{-1}\left(  \mathbf{Y}-\boldsymbol{\Theta}\right)  ^{T}\boldsymbol{\Theta}+k^{-1}%
\boldsymbol{\Theta}^{T}\left(  \mathbf{Y}-\boldsymbol{\Theta}\right)  \\
& +k^{-1}\left(  \mathbf{Y}-\boldsymbol{\Theta}\right)  ^{T}\left(  \mathbf{Y}%
-\boldsymbol{\Theta}\right)
\end{align*}
as $\mathbf{A}_{1}$, $\mathbf{A}_{2}$ and $\mathbf{A}_{2}^{T}$ and
$\mathbf{A}_{3}$, we see that $\lim\limits_{k\rightarrow\infty}\mathbf{A}%
_{1}=\mathbf{H}$ by assumption (\ref{2}). We claim that $\lim\limits_{k\rightarrow
\infty}\left\Vert \mathbf{A}_{2}\right\Vert =0=\lim_{k\rightarrow\infty
}\left\Vert \mathbf{A}_{2}^{T}\right\Vert $ a.s. The $\left(
i,j\right)  $ entry $a_{2,ij}$ of $\mathbf{A}_{2}$ is $a_{2,ij}=k^{-1}%
\sum\limits_{l=1}^{k}\tilde{a}_{2,ij,l}$, where $\tilde{a}_{2,ij,l}=\left(
y_{li}-\theta_{li}\right)  \theta_{lj}$ for $1\leq l\leq k$. Clearly, $\tilde{\mathbb{E}}\left[
\tilde{a}_{2,ij,l}\right]  =0$. Set $\tilde{y}_{li}=y_{li}-\theta_{li}$. By
Lemma \ref{Lm:UnifConv}, uniformly bounded $4$th conditional moments of $y_{ij}$ and H\"{o}lder's
inequality,%
\[
\sup_{k\geq1}\max_{1\leq l\leq k}\tilde{\mathbb{V}}\left[  \tilde{a}_{2,ij,l}\right]
\leq C\sup_{k\geq1}\max_{1\leq l\leq k}\left(  \tilde{\mathbb{E}}\left[  \tilde
{y}_{li}^{4}\right]  \right)  ^{1/2}\leq C\text{.}%
\]
By independence conditional $\mathbf{M}$,%
\[
\sum_{k\geq1}\dfrac{1}{k}\tilde{\mathbb{V}}\left[  a_{2,ij}\right]  \leq C\sum
_{k\geq1}\dfrac{1}{k^{3}}\sum_{l=1}^{k}\left( \tilde{ \mathbb{E}}\left[  \tilde
{y}_{li}^{4}\right]  \right)  ^{1/2}\leq C\sum_{k\geq1}\dfrac{1}{k^{2}}%
=\dfrac{C\pi^{2}}{6}\text{.}%
\]
Therefore, Theorem 1 of \cite{Walk:2005} implies that $\lim\limits_{k\rightarrow
\infty}a_{2,ij}=0$ a.s., which validates the claim.

Consider the last term $\mathbf{A}_{3}$, whose $\left(  i,j\right)  $th
off-diagonal entry $a_{3,ij}$ can be written as $a_{3,ij}=k^{-1}\sum\limits_{l=1}%
^{k}\tilde{a}_{3,ij,l}$, where $\tilde{a}_{3,ij,l}=\left(  y_{li}%
-\theta_{li}\right)  \left(  y_{lj}-\theta_{lj}\right)  $. Clearly, $\tilde{\mathbb{E}}\left[
\tilde{a}_{3,ij,l}\right]  =0$. The same reasoning as above implies%
\[
\sum_{k\geq1}\dfrac{1}{k}\tilde{\mathbb{V}}\left[  a_{3,ij}\right]  \leq C\sum
_{k\geq1}\dfrac{1}{k^{3}}\sum_{l=1}^{k}\left(  \tilde{\mathbb{E}}\left[  \tilde
{y}_{li}^{4}\right]  \right)  ^{1/2}\left(  \tilde{\mathbb{E}}\left[  \tilde{y}%
_{lj}^{4}\right]  \right)  ^{1/2}\leq C\sum_{k\geq1}\dfrac{1}{k^{2}}%
=\dfrac{C\pi^{2}}{6}%
\]
when $i\neq j$. So, Theorem 1 of \cite{Walk:2005} implies that
$\lim\limits_{k\rightarrow\infty}a_{3,ij}=0$ a.s. The diagonal entries of
$\mathbf{A}_{3}$ can be written as $a_{3,ii}=k^{-1}\sum\limits_{l=1}^{k}\tilde
{a}_{3,ii,l}$, where $\tilde{a}_{3,ii,l}=\left(  y_{li}-\theta_{li}\right)  ^{2}$.
Since $\tilde{\mathbb{E}}\left[  \tilde{a}_{3,ii,l}\right]  =\delta_{li}$ and
\[
\tilde{\mathbb{V}}\left[  \tilde{a}_{3,ii,l}\right]  \leq\delta_{i}^{2}+\tilde{\mathbb{E}}%
\left[  \tilde{y}_{li}^{4}\right]  \leq2\tilde{\mathbb{E}}\left[  \tilde{y}_{li}%
^{4}\right]  \leq C\text{,}%
\]
it follows that%
\[
\sum_{k\geq1}\dfrac{1}{k}\tilde{\mathbb{V}}\left[  a_{3,ii}\right]  \leq\sum_{k\geq
1}\dfrac{C}{k^{2}}=\dfrac{C\pi^{2}}{6}\text{.}%
\]
Hence, Theorem 1 of \cite{Walk:2005} implies $\lim\limits_{k\rightarrow\infty
}\left\vert a_{3,ii}-\bar{\delta}_{ki}\right\vert =0$ a.s.\ and
$\lim\limits_{k\rightarrow\infty}\left\Vert \mathbf{A}_{3}-\mathbf{D}_{k}\right\Vert
=0$. Combining the limiting terms obtained above completes the proof.

\subsection{Proof of Corollary \ref{Lm:ConvEigensys}}

We show (\ref{1a}) first. By (\ref{6}) and
Wielandt-Hoffman (WH) inequality of \cite{Hoffman:1953}, we immediately have (\ref{1a}), i.e.,
\begin{equation}
\lim_{k\rightarrow\infty}\max_{1\leq i\leq n}\left\vert \beta_{k,i}
-\alpha_{i}\right\vert =0\text{ a.s.}.\label{15}
\end{equation}

Now we show the rest of the assertions.
Let $\left\{  \lambda_{l}\right\}  _{l=1}^{s}$ with $1\leq s\leq n$ be the
distinct eigenvalues of $\mathbf{H}$ ordered into $\lambda_{l}>\lambda_{l+1}$
for $1\leq l\leq s-1$. Note that $s_l=0$ since the rank of $\mathbf{H}$ is $r$ and $\mathbf{H}$ is positive semi-definite.
Let $\varepsilon_{0}=\min\limits_{1\leq l\leq s_{0}}\left\{  \lambda_{l}%
-\lambda_{l+1}\right\}  $ and pick $\varepsilon>0$ such that $\varepsilon
<4^{-1}\varepsilon_{0}$, where $s_{0}=\max\left\{
l:\lambda_{l}>0\right\} \leq r <n $. From (\ref{15}), we see that there exists some
$k_{0}\in\mathbb{N}$ and a partition of
$\left\{  1,...,n\right\}  $ into its subsets $\left\{  A_{l}\right\}
_{l=1}^{s}$ for which, whenever $k\geq k_{0}$, $\max\limits_{1\leq l\leq s}\max\limits_{i\in A_{l}}\left\vert
\beta_{k,i}-\lambda_{l}\right\vert < 4^{-1}\varepsilon$ but%
\[
\min_{1\leq l<l^{\prime}\leq s}\min\left\{  \left\vert \beta_{k,i}-\beta
_{k,j}\right\vert :i\in A_{l},j\in A_{l^{\prime}},l\neq l^{\prime}\right\}
>4^{-1}\varepsilon_{0}\text{.}%
\]
Namely, $\left\{  \beta_{k,i}\right\}  _{i=1}^{n}$ separate into $s$ groups
$\left\{  \beta_{k,i}:i\in A_{l}\right\}  $ for $1\leq l\leq s$ each with
center $\lambda_{l}$ but diameter at most $2^{-1}\varepsilon$.
Therefore, $\lim_{k \rightarrow \infty}\left\{\beta_{k,i}:i=1,\ldots,n\right\}=\left\{\lambda_l:l=1,\ldots,s\right\}$
a.s..

Since $\mathbf{W}>0$, $\rank\left(  \mathbf{M}\right)
=r$ and $\mathbf{H}$ is symmetric, $\mathbf{H}$ is diagonalizable with
$\rank\left(  \mathbf{H}\right)  =r$ and the geometric multiplicity (gm)
of each $\lambda_{l}$ equals its algebraic multiplicity (am).
Therefore, the linear space spanned by the union of
the eigenvectors corresponding to all non-zero eigenvalues of $\mathbf{H}$
must be $\left\langle \left\{  \mathbf{v}%
_{i}\right\}  _{i=1}^{r}\right\rangle $ and $\rank\left(
\left\langle \left\{  \mathbf{v}%
_{i}\right\}  _{i=1}^{r}\right\rangle\right)  =r$ must hold,
where $\mathbf{v}_{i}$ is an eigenvector corresponding to some $\alpha_{j}$
for $1\leq j\leq r$. Moreover, $\left\langle \left\{  \mathbf{v}%
_{i}\right\}  _{i=1}^{r}\right\rangle =\Pi_{\mathbf{M}}$. Fix an $l$ and any $\boldsymbol{\gamma}%
\in\mathbb{R}^{n}$ with $\left\Vert \boldsymbol{\gamma}\right\Vert <\infty$.
Let $\mathbf{u}_{k,i}$ be any eigenvector that is part of the basis for the
eigenspace $U_{k,i}$ of $\beta_{k,i}$. Then, from (\ref{6}) and (\ref{1a}), we
see that
\[
b_{k}\left(  \boldsymbol{\gamma}\right)  =\left(  \mathbf{R}_{k}-\beta
_{k,i}\mathbf{I}\right)  \boldsymbol{\gamma}-\left(  \mathbf{R}_{k}%
-\beta_{k,j}\mathbf{I}\right)  \boldsymbol{\gamma}%
\]
satisfies $\lim\limits_{k\rightarrow\infty}\left\Vert b_{k}\left(  \boldsymbol{\gamma
}\right)  \right\Vert =0$ a.s.\ for all $i,j\in A_{l}$. Therefore,
$\lim\limits_{k\rightarrow\infty}U_{k,i}\triangle U_{k,j}=\varnothing$ a.s.\ for all
$i,j\in A_{l}$ and asymptotically there are only $s$ linearly independent
eigenspaces $\left\{  U_{k,i_{l}}\right\}  _{l=1}^{s}$ with $1\leq i_{l}\leq
n$ corresponding to $\left\{
\lim\limits_{k\rightarrow\infty}\beta_{k,i}:i\in A_{l},1\leq l\leq
s\right\}  $. On the other hand, for any $i \in A_{l}$,%
\[
a_{k}\left(  \boldsymbol{\gamma}\right)  =\left(  \mathbf{R}_{k}-\beta
_{k,i}\mathbf{I}\right)  \boldsymbol{\gamma}-\left(  \mathbf{H}-\lambda
_{l}\mathbf{I}\right)  \boldsymbol{\gamma}%
\]
satisfies $\lim\limits_{k\rightarrow\infty}\left\Vert a_{k}\left(  \boldsymbol{\gamma
}\right)  \right\Vert =0$ a.s.. Let $\mathbf{v}$ be
any eigenvector that is part of the basis for the eigenspace $V_{l}$ of
$\lambda_{l}$. Then, for any $i\in A_{l}$,%
\[
\left\Vert \left(  \mathbf{R}_{k}-\beta_{k,i}\mathbf{I}\right)  \mathbf{v}%
\right\Vert \leq\left\Vert \left(  \mathbf{H}-\lambda_{l}\mathbf{I}\right)
\mathbf{v}\right\Vert +\left\Vert a_{k}\left(  \boldsymbol{\gamma}\right)
\right\Vert \rightarrow 0 \text{\ a.s.}
\]
and%
\[
\left\Vert \left(  \mathbf{H}-\lambda_{l}\mathbf{I}\right)  \mathbf{u}%
_{k,i}\right\Vert \leq\left\Vert \left(  \mathbf{R}_{k}-\beta_{k,i}%
\mathbf{I}\right)  \mathbf{u}_{k,i}\right\Vert +\left\Vert a_{k}\left(
\boldsymbol{\gamma}\right)  \right\Vert \rightarrow 0 \text{\ a.s.}
\]
since $\left\Vert \mathbf{v}\right\Vert =1$ and $\Pr\left(  \left\Vert
\mathbf{u}_{k,i}\right\Vert <\infty\right)  =1$. Consequently, $V_{l}%
\subseteq\lim\limits_{k\rightarrow\infty}U_{k,i}$ and $\lim\limits_{m\rightarrow\infty
}U_{k,i}\subseteq V_{l}$ a.s. for any $i\in A_{l}$. Since we have already
shown that $\left\{  U_{k,i}\right\}  _{i=1}^{n}$ asymptotically reduces to
$\left\{  U_{k,i_{l}}\right\}  _{l=1}^{s}$ corresponding to $\left\{
\beta_{k,i_{l}}\right\}  _{l=1}^{s}\subseteq\left\{  \beta_{k,i}\right\}
_{i=1}^{n}$ for which $\lim\limits_{k\rightarrow\infty}\max\limits_{1\leq l\leq s}\left\vert
\beta_{k,i_{l}}-\lambda_{l}\right\vert =0$ a.s., we see that all eigenvectors
corresponding to the $r$ largest eigenvalues of $\mathbf{R}_{k}$ together
asymptotically spans $\left\langle \left\{  \mathbf{v}%
_{i}\right\}  _{i=1}^{r}\right\rangle $ a.s..
This yields $\lim\limits_{k\rightarrow\infty}\left\vert S\right\vert =r$ a.s. and
(\ref{3a}). The proof is completed.

\subsection{Proof of Proposition \ref{Prop:CLTRandomPart}}

We remark that our proof of Proposition \ref{Prop:CLTRandomPart} follows a similar strategy
given in \cite{Leek:2011} but uses slightly different techniques.

We show the first claim. Recall $\boldsymbol{\Theta}=\mathbf{\Phi M}$. It suffices to show that each of $k^{-1}%
\boldsymbol{\Theta}^{T}\mathbf{E}$ and $k^{-1}\mathbf{E}^{T}\mathbf{E}$ is a linear
mapping of $\sum\limits_{i=1}^{k}\mathbf{z}_{i}$ with Lipschitz constant $1$ as
follows. The $\left(  i,j\right)  $th entry of $k^{-1}\mathbf{E}^{T}%
\mathbf{E}$ is
\[
k^{-1}\sum\limits_{l=1}^{k}e_{li}e_{lj}=k^{-1}\sum\limits_{l=1}%
^{k}\left\langle \mathbf{q}_{j},e_{li}\mathbf{e}_{l}\right\rangle
=\left\langle \mathbf{q}_{j},k^{-1}\sum\limits_{l=1}^{k}e_{li}\mathbf{e}%
_{l}\right\rangle,
\]
where $\mathbf{q}_{i}=\left(  0,...,0,1,0,...,0\right)  $ (i.e., only the
$i$th entry is $1$; others are $0$) and $\left\langle \cdot,\cdot\right\rangle
$ is the inner product in Euclidean space. In other words, the $\left(
i,j\right)  $th entry of $k^{-1}\mathbf{E}^{T}\mathbf{E}$ is exactly the
$\left(  in+j\right)  $th entry of $\sum\limits_{i=1}^{k}\mathbf{z}_{i}$. Further,
the $\left(  i,j\right)  $th entry of $k^{-1}\boldsymbol{\Theta}^{T}\mathbf{E}$ is%
\[
k^{-1}\sum\limits_{l=1}^{k}\boldsymbol{\phi}_{l}\mathbf{m}^{i}e_{lj}%
=k^{-1}\sum\limits_{l=1}^{k}\left\langle \mathbf{q}_{j},\boldsymbol{\phi
}_{l}\mathbf{m}^{i}\mathbf{e}_{l}\right\rangle =\left\langle \mathbf{q}%
_{j},k^{-1}\sum\limits_{l=1}^{k}\boldsymbol{\phi}_{l}\mathbf{m}%
^{i}\mathbf{e}_{l}\right\rangle \text{,}%
\]
i.e., the $(n^{2}+ni+j)$th entry of $\sum\limits_{i=1}^{k}\mathbf{z}_{i}$.

Notice that linearity of picking an entry as a mapping described above is
invariant under matrix transpose, we see that $k^{-1}\mathbf{E}%
^{T}\boldsymbol{\Theta}$ is also a linear mapping
of $\sum\limits_{i=1}^{k}\mathbf{z}_{i}$ with Lipschitz constant $1$. Consequently,
$\mathbf{F}$ is a linear mapping of $\left\{
\mathbf{z}_{i}\right\}  _{i=1}^{k}$ with Lipschitz constant $1$.

Now we show the second claim. Set $\sum\limits_{i=1}^{k}\mathbf{\tilde{z}}_{i}$ with $\mathbf{\tilde{z}%
}_{i}=\sqrt{k}\left(  \mathbf{z}_{i}-\tilde{\mathbb{E}}\left[  \mathbf{z}_{i}\right]
\right)  $. We will verify that $\sum\limits_{i=1}^{k}\mathbf{\tilde{z}}_{i}$ is
asymptotically Normally distributed. Clearly,\ among the long vector
$\tilde{\mathbb{E}}\left[  \mathbf{z}_{i}\right]  $, only the $j$th entry of
$\tilde{\mathbb{E}}\left[  e_{ij}\mathbf{e}_{i}\right]  $ is nonzero and it is
$k^{-1}\delta_{ij}$, which means that $\tilde{\mathbb{E}}\left[  \mathbf{z}_{i}\right]  $
has an easy form. By the multivariate central limit theorem (MCLT), e.g.,
see page 20 of \cite{Vaart:1998}, to show the asymptotic normality of
$\sum\limits_{i=1}^{k}\mathbf{\tilde{z}}_{i}$ as $k\rightarrow\infty$, it suffices to
show
\begin{equation}
v_{k}^{\varepsilon}=\sum_{i=1}^{k}\tilde{\mathbb{E}}\left[  \left\Vert \mathbf{\tilde
{z}}_{i}\right\Vert ^{2}1_{\left\{  \left\Vert \mathbf{\tilde{z}}%
_{i}\right\Vert >\varepsilon\right\}  }\right]  \rightarrow0\text{ as
}k\rightarrow\infty\label{81a}%
\end{equation}
for any $\varepsilon>0$ and
\begin{equation}
\sum_{i=1}^{k}\widetilde{Cov}\left[  \mathbf{\tilde{z}}_{i}\right]  \rightarrow
\boldsymbol{\Sigma}_{\infty}\text{ as }k\rightarrow\infty\label{82a}%
\end{equation}
for some matrix $\boldsymbol{\Sigma}_{\infty}$ in order that $\sum\limits_{i=1}%
^{k}\mathbf{\tilde{z}}_{i}\rightsquigarrow N\left(  \mathbf{0}%
,\boldsymbol{\Sigma}_{\infty}\right)  $, where
$\widetilde{Cov}$ is the conditional covariance operator given $\mathbf{M}$ and $\rightsquigarrow$ denotes weak convergence.

We verify (\ref{81a}) first. Since $\sup\limits_{k}\max\limits_{i,j}\tilde{\mathbb{E}}\left[
y_{ij}^{8}\right]  \leq C$, $\sup\limits_{k}\max\limits_{i}\left\Vert \boldsymbol{\phi}%
_{i}\right\Vert <\infty$ and $n$ is finite, the identity%
\begin{equation}
k\left\Vert \mathbf{\tilde{z}}_{i}\right\Vert ^{2}=\sum_{j\neq l;1\leq j,l\leq
n}e_{il}^{2}e_{ij}^{2}+\sum_{j=1}^{n}\left(  e_{ij}^{2}-\delta_{ij}\right)
^{2}+\sum_{j=1}^{n}\sum_{l=1}^{n}\left(  \boldsymbol{\phi}_{i}\mathbf{m}%
^{j}e_{il}\right)  ^{2}\label{14}%
\end{equation}
for each $1\leq i\leq k$ implies, via H\"{o}lder's inequality,
\[
\tilde{\mathbb{E}}\left[  \left\Vert \mathbf{\tilde{z}}_{i}\right\Vert ^{2}1_{\left\{
\left\Vert \mathbf{\tilde{z}}_{i}\right\Vert >\varepsilon\right\}  }\right]
\leq k^{-1}\left(  \tilde{\mathbb{E}}\left[  \vartheta_{i}^{2}\right]  \Pr\left(
\vartheta_{i}>k\varepsilon^{2}\right)  \right)  ^{1/2}\leq\dfrac{C}%
{k^{3/2}\varepsilon}%
\]
and $v_{k}^{\varepsilon}\leq Ck^{-1/2}\varepsilon^{-1}\rightarrow0$ as
$k\rightarrow\infty$, where $\vartheta_{i}$ denotes the right hand side (RHS)
of (\ref{14}). Hence, (\ref{81a}) is valid.

It is left to verify (\ref{82a}). Let $\boldsymbol{\Sigma}_{_{i}}^{\ast
}=\widetilde{Cov}\left(  \mathbf{\tilde{z}}_{i}\right)  $. Then
$k\boldsymbol{\Sigma}_{_{i}}^{\ast}$ has entries zero except at the blocks
$\mathbf{S}_{i1,jl}=\widetilde{Cov}\left(  e_{ij}\mathbf{e}_{i},e_{il}\mathbf{e}%
_{i}\right)  $, $\mathbf{S}_{i2,jl}=\widetilde{Cov}\left(  \boldsymbol{\phi}_{i}%
\mathbf{m}^{j}\mathbf{e}_{i},\boldsymbol{\phi}_{i}\mathbf{m}^{l}\mathbf{e}%
_{i}\right)  $ and $\mathbf{S}_{i4,jl}=\widetilde{Cov}\left(  e_{ij}\mathbf{e}%
_{i},\boldsymbol{\phi}_{i}\mathbf{m}^{l}\mathbf{e}_{i}\right)  $.
Specifically, $\mathbf{S}_{i1,jl}\left(  r,r^{\prime}\right)  =\tilde{\mathbb{E}}%
\left(  e_{ir}^{4}\right)  $ if $r=r^{\prime}=j=l$ and is $0$ otherwise;
$\mathbf{S}_{i2,jl}=\boldsymbol{\phi}_{i}\mathbf{m}^{j}\boldsymbol{\phi}%
_{i}\mathbf{m}^{l}\mathbf{\Delta}_i$ with $\mathbf{\Delta}_i=\diag\left\{  \delta_{i1},...,\delta
_{in}\right\}  $; $\mathbf{S}_{i4,jl}\left(
r,r^{\prime}\right)  =\boldsymbol{\phi}_{i}\mathbf{m}^{l}\tilde{\mathbb{E}}\left[
e_{ir}^{3}\right]  $ if $r=r^{\prime}=j$ and is $0$ otherwise. By the joint independence among
$\left\{  \mathbf{\tilde{z}}_{i}\right\}  _{i=1}^{k}$, we have $\widetilde{Cov}\left[
\sum\limits_{i=1}^{k}\mathbf{\tilde{z}}_{i}\right]  =
\boldsymbol{\Sigma}_{k}^{\ast}=\sum\limits_{i=1}^{k}\boldsymbol{\Sigma}_{_{i}}%
^{\ast}$. Further, $\boldsymbol{\Sigma}_{k}^{\ast}\rightarrow\boldsymbol{\Sigma
}_{\infty}$ as $k\rightarrow\infty$ when, for each $1\leq j,l\leq n$,

\begin{enumerate}
\item $k^{-1}\sum\limits_{i=1}^{k}\tilde{\mathbb{E}}\left[  e_{ij}^{4}\right]  $ and
$k^{-1}\sum\limits_{i=1}^{k}\boldsymbol{\phi}_{i}\tilde{\mathbb{E}}\left[  e_{ij}^{3}\right]
$\ are convergent (so that both $k^{-1}\sum\limits_{i=1}^{k}\mathbf{S}_{i1,jl}$ and
$k^{-1}\sum\limits_{i=1}^{k}\mathbf{S}_{i4,jl}$ converge).

\item $k^{-1}\sum\limits_{i=1}^{k}\mathbf{S}_{i2,jl}$ is convergent.
\end{enumerate}
By assumption A3), the above needed convergence is ensured, meaning that
$\boldsymbol{\Sigma}_{k}^{\ast}$ $\rightarrow\boldsymbol{\Sigma}_{\infty}$ for
some $\boldsymbol{\Sigma}_{\infty}\geq0$, i.e., (\ref{82a}) holds. Therefore,
$\sum\limits_{i=1}^{k}\mathbf{\tilde{z}}_{i}$ is asymptotically Normally distributed
as $N\left(  \mathbf{0},\boldsymbol{\Sigma}_{\infty}\right)  $. The proof is completed.

\subsection{Proof of \autoref{Prop:EstRank}}

We will use the notations in the proof of Proposition \ref{Prop:CLTRandomPart} and aim
to show $\left\Vert \mathbf{\hat{R}}_{k}-\mathbf{H}\right\Vert =O_{\Pr}\left(
\tau_{k}\right)  $, which then by Wielandt-Hoffman (WH) inequality of
\cite{Hoffman:1953} implies $\left\Vert \boldsymbol{\hat{\alpha}}%
_{k}-\boldsymbol{\alpha}\right\Vert =O_{\Pr}\left(  \tau_{k}\right)  $, where
$\boldsymbol{\hat{\alpha}}_{k}=\left(  \hat{\alpha}_{k,1},...,\hat{\alpha
}_{k,n}\right)  $ and $\boldsymbol{\alpha}=\left(  \alpha_{1},...,\alpha
_{n}\right)  $. Recall $k^{-1}\mathbf{Y}^{T}\mathbf{Y}=\mathbf{F}%
+k^{-1}\boldsymbol{\Theta}^{T}\boldsymbol{\Theta}$ and $\mathbf{F}=f_{0}\left(  \sum\limits_{i=1}%
^{k}\mathbf{z}_{i}\right)  $ by Proposition \ref{Prop:CLTRandomPart}, where $f_{0}$ is
linear with Lipschitz constant $1$. We see that the linearity of $f_{0}$ and
that the only nonzero entries of $\tilde{\mathbb{E}}\left[  \mathbf{z}_{i}\right]  $
are $k^{-1}\delta_{ij}$ at indices $\left(  j-1\right)  n+j$ for $1\leq j\leq n$
together imply%
\[
\sqrt{k}\mathbf{F}=f_{0}\left(  \sum\limits_{i=1}^{k}\sqrt{k}\left(
\mathbf{z}_{i}-\tilde{\mathbb{E}}\left[  \mathbf{z}_{i}\right]  \right)  \right)
+\sqrt{k}\mathbf{D}_{k}=f_{0}\left(  \sum\limits_{i=1}^{k}\mathbf{\tilde{z}%
}_{i}\right)  +\sqrt{k}\mathbf{D}_{k}\text{.}%
\]
Therefore, the Lipschitz property of $f_{0}$ and the asymptotic normality of
$\sum\limits_{i=1}^{k}\mathbf{\tilde{z}}_{i}$ force%
\begin{align*}
\sqrt{k}\left(  \mathbf{\hat{R}}_{k}-\mathbf{H}\right)    & =f_{0}\left(
\sum\limits_{i=1}^{k}\mathbf{\tilde{z}}_{i}\right)  +\sqrt{k}\left(
\mathbf{\hat{D}}_{k}-\mathbf{D}_{k}\right)  +\sqrt{k}\left(  \mathbf{M}%
^{T}\left(  k^{-1}\boldsymbol{\Phi}^{T}\boldsymbol{\Phi}-\mathbf{W}\right)
\mathbf{M}\right)  \\
& =O_{\Pr}\left(  1\right)  +\sqrt{k}\varepsilon_{k}+\sqrt{k}O\left(  c_{k}\right)
\text{,}%
\end{align*}
where we have used assumptions (\ref{11}) and (\ref{2}). Hence, (\ref{26})
holds with $\tau_{k}$ set by (\ref{25}). Finally, we show (\ref{20}) and (\ref{eq:rankest}). From
(\ref{26}), we see that $\max\left\{  \alpha_{i}-\tilde{c}_{i}\tau_{k},0\right\}
\leq\hat{\alpha}_{k,i}\leq\alpha_{i}+\tilde{c}_{i}\tau_{k}$ for
some finite constant $\tilde{c}_{i} \geq 0$ for each $1\leq i\leq n$. Since $\alpha_{r}>0$
but $\alpha_{r+1}=0$, we see that $\tilde{\tau}_{k}^{-1}\hat{\alpha}_{k,i}\geq\tilde
{\tau}_{k}^{-1}\alpha_{r}-\tilde{c}_{i}\tilde{\tau}_{k}^{-1}\tau_{k}\rightarrow\infty$
for $1\leq i\leq r$ but $\tilde{\tau}_{k}^{-1}\hat{\alpha}_{k,i}\leq
\tilde{c}_{i}\tilde{\tau}_{k}^{-1}\tau_{k}\rightarrow0$ for $r+1\leq i\leq n$.
So, (\ref{20}) and (\ref{eq:rankest}) hold. The proof is completed.

\subsection{Proof of Lemma \ref{Lm:AveVar}}

Let $w_{ij}=v\left(  y_{ij}\right)  $. Then $w_{ij}$ are mutually independent
and
\[
\sum_{k\geq1}k^{-1}\tilde{\mathbb{V}}\left[  \hat{\delta}_{kj}\right]  =\sum_{k\geq
1}k^{-2}\max_{1\leq i\leq k}\tilde{\mathbb{V}}\left[  w_{ij}\right]  \leq C\sum
_{k\geq1}k^{-2}<\infty\text{.}%
\]
Therefore, Theorem 1 of \cite{Walk:2005} implies that $\tilde{d}_{kj}%
=\hat{\delta}_{kj}-\bar{\delta}_{kj}=\hat{\delta}_{kj}-\tilde{\mathbb{E}}\left[
\hat{\delta}_{kj}\right]  \rightarrow0$ a.s. as $k\rightarrow\infty$. Now we
show the second claim. Let $\tilde{y}_{ij}=v\left(  y_{ij}\right)
-\delta_{ij}$ and $b_{ij}=k^{-1/2}\tilde{y}_{ij}$. Then $\sum\limits_{l=1}^{k}%
b_{lj}=\sqrt{k}\tilde{d}_{kj}$. For any $\varepsilon>0$, assumption (\ref{29})
implies%
\[
\tilde{a}_{i}=\tilde{\mathbb{E}}\left[  b_{lj}^{2}1_{\left\{  \left\vert
b_{lj}\right\vert >\varepsilon\right\}  }\right]  \leq k^{-1}\left(
\tilde{\mathbb{E}}\left[  \tilde{y}_{ij}^{4}\right]  \Pr\left(  \left\vert \tilde
{y}_{ij}\right\vert >k^{1/2}\varepsilon\right)  \right)  ^{1/2}\leq\dfrac
{C}{k^{5/4}\varepsilon}%
\]
and
\[
\lim_{k\rightarrow\infty}\sum_{i=1}^{k}\tilde{a}_{i}\leq\lim_{k\rightarrow
\infty}\dfrac{kC}{k^{5/4}\varepsilon}=0\text{.}%
\]
By assumption (\ref{18}),
$\lim\limits_{k\rightarrow\infty}\sum\limits_{i=1}^{k}\tilde{\mathbb{V}}\left[  b_{ij}\right]=\sigma_{j}$ %
for each $1\leq j\leq n$. Thus, the conditions of MCLT (e.g., see \citealp{Vaart:1998}) are satisfied and
$\sum\limits_{l=1}^{k}b_{lj}$ converges in distribution to a Normal random variable with mean
$0$ and variance $\sigma_{j}$. The proof is completed.

\clearpage
\setcounter{table}{0}
\renewcommand{\thetable}{S\arabic{table}}
\section{Performance of $\hat{r}$ in estimating $r$ \label{AppC:ResSimu}}
In the following tables, we provide an assessment of the estimator $\hat{r}$ in several scenarios that extend beyond that shown in \autoref{TbA1}.  Data were simulated under model \eqref{7} over a range of $n$ and $r$ values under the five distributions listed.  Shown is the number of times that $\hat{r}=r$ among $100$ simulated data sets for each scenario.  Also shown in parentheses is the number of times that $\hat{r} < r$, if any instances occurred.

\begin{table}[H]
\centering
\caption{Performance of $\hat{r} $ as an estimator of $r$ when $n=15$ and $r \in \left\{ 1,2,3,4\right \}$.} \label{tab:a1}
\begin{tabular}{|l|l|l|l|l|l|l|}
  \hline
      $k$     & Binomial & Gamma & NegBin & Normal & Poisson \\
  \hline
 \multicolumn{1}{|c}{\ } &  \multicolumn{5}{c|}{$n=15$ and $r=1$} \\
\hline
       1000  &             96 &    82 &    82 &    100 &      91 \\
      5000  &       99 &    89 &   86 &    100 &      92 \\
      10,000  &     96 &    76 &    91 &     99 &      93 \\
      100,000  &       100 &    90 &    86 &     99 &      94 \\
   \hline
 \multicolumn{1}{|c}{\ } &  \multicolumn{5}{c|}{$n=15$ and $r=2$} \\
\hline
       1000  &      96 &    92 &                5 (85) &    100 &      94 \\
      5000  &       96 &    90 &               93 &     98 &      97 \\
      10,000  &       97 &    93 &               90 &     96 &      95 \\
      100,000  &       99 &    96 &               93 &     99 &      97 \\
   \hline
 \multicolumn{1}{|c}{\ } &  \multicolumn{5}{c|}{$n=15$ and $r=3$} \\
\hline
       1000  &      95 &    27 (68) &         36 (51) &     96 &      94 \\
      5000  &       97 &    44 (46) &         45 (46) &     99 &      93 \\
      10,000  &       100 &    89 &               76 (20) &     96 &      95 \\
      100,000  &       97 &    95 &               90 &     98 &      96 \\
   \hline
 \multicolumn{1}{|c}{\ } &  \multicolumn{5}{c|}{$n=15$ and $r=4$} \\
\hline
       1000  &      51 (48) &    25 (29) &     9 (2) &    100 &      94 \\
      5000  &       99 &    20 (72) &          49 (5) &     97 &      94 \\
      10,000  &       100 &    60 (27) &           28 (35) &     97 &      93 \\
      100,000  &       98 &    91 &               90 &     97 &      93 \\
   \hline
\end{tabular}
\end{table}

\clearpage
\begin{table}[H]
\centering
\caption{Performance of $\hat{r} $ as an estimator of $r$ when $n=100$ and $r \in \left\{ 1,2,3,4,5\right \}$.} \label{tab:a2}
\begin{tabular}{|l|l|l|l|l|l|l|}
  \hline
      $k$     & Binomial & Gamma & NegBin & Normal & Poisson \\
  \hline
 \multicolumn{1}{|c}{\ } &  \multicolumn{5}{c|}{$n=100$ and $r=1$} \\
\hline
       1000  &             100 &     99 (1) &       99 &        100 &      100 \\
       5000  &             100 &     100 &        100 &          100 &      100 \\
       10,000 &              100 &        100&        100 &     100 &      100 \\
       100,000 &              100 &        100 &            100 &        100 &      100 \\
  \hline
 \multicolumn{1}{|c}{\ } &  \multicolumn{5}{c|}{$n=100$ and $r=2$} \\
\hline
       1000  &             0 (100) &     100 &       100 &        100 &      100 \\
       5000  &             0 (100) &     100 &        100 &          100 &      100 \\
       10,000 &              0 (100) &        100&        100 &     100 &      100 \\
       100,000 &              0 (100) &        100 &            100 &        100 &      100 \\
   \hline
 \multicolumn{1}{|c}{\ } &  \multicolumn{5}{c|}{$n=100$ and $r=3$} \\
\hline
       1000  &             0 (100) &     100 &       100 &        100 &      100 \\
       5000  &             0 (100) &     100 &        100 &          100 &      100 \\
       10,000 &              0 (100) &        100&        100 &     100 &      100 \\
       100,000 &              0 (100) &        100 &            100 &        100 &      100 \\
   \hline
 \multicolumn{1}{|c}{\ } &  \multicolumn{5}{c|}{$n=100$ and $r=4$} \\
\hline
       1000  &             0 (100) &     77 (23) &       24 (76) &        100 &      100 \\
       5000  &             0 (100) &     100 &        100 &          100 &      100 \\
       10,000 &              0 (100) &        100&        100 &     100 &      100 \\
       100,000 &              0 (100) &        100 &            100 &        100 &      100 \\
   \hline
 \multicolumn{1}{|c}{\ } &  \multicolumn{5}{c|}{$n=100$ and $r=5$} \\
\hline
       1000  &             0 (100) &     35 (52) &       40 &        100 &      100 \\
       5000  &             0 (100) &     96 (4) &        100 &          100 &      100 \\
       10,000 &              0 (100) &        100&        100 &     100 &      100 \\
       100,000 &              0 (100) &        100 &            100 &        100 &      100 \\
   \hline
\end{tabular}
\label{Tbn100r5}
\end{table}

\clearpage
\begin{table}[H]
\centering
\caption{Performance of $\hat{r} $ as an estimator of $r$ when $n=200$ and $r \in \left\{6,8,10,12 \right \}$.} \label{tab:a3}
\begin{tabular}{|l|l|l|l|l|l|l|}
  \hline
      $k$     & Binomial & Gamma & NegBin & Normal & Poisson \\
  \hline
 \multicolumn{1}{|c}{\ } &  \multicolumn{5}{c|}{$n=200$ and $r=6$} \\
\hline
       1000  &             0 (100) &     59 (10) &       1 &        100 &      100 \\
       5000  &             0 (100) &     100 &        100 &          100 &      100 \\
       10,000 &              0 (100) &        100&        100 &     100 &      100 \\
       100,000 &              0 (100) &        100 &            100 &        100 &      100 \\
   \hline
 \multicolumn{1}{|c}{\ } &  \multicolumn{5}{c|}{$n=200$ and $r=8$} \\
\hline
       1000  &             0 (100) &     0 &       0 &        100 &      100 \\
       5000  &             0 (100) &     0 &        0 &          100 &      100 \\
       10,000 &              0 (100) &        78&        0 &     100 &      100 \\
       100,000 &              0 (100) &        100 &            100 &        100 &      100 \\
   \hline
 \multicolumn{1}{|c}{\ } &  \multicolumn{5}{c|}{$n=200$ and $r=10$} \\
\hline
       1000  &             0 (100) &     0 &       0 &        100 &      100 \\
       5000  &             0 (100) &     0 &        0 &          100 &      100 \\
       10,000 &              0 (100) &        0&        0 &     100 &      100 \\
       100,000 &              0 (100) &        69 &            11 &        100 &      100 \\
   \hline
 \multicolumn{1}{|c}{\ } &  \multicolumn{5}{c|}{$n=200$ and $r=12$} \\
\hline
       1000  &             0 (100) &     0 &       0 &        100 &      100 \\
       5000  &             0 (100) &     0 &        0 &          100 &      100 \\
       10,000 &              0 (100) &        0&        0 &     100 &      100 \\
       100,000 &              0 (100) &        0 &            0 &        100 &      100 \\
   \hline
\end{tabular}
\end{table}

\clearpage
\setcounter{figure}{0}
\renewcommand{\thefigure}{S\arabic{figure}}
\section{Performance of Estimator of $\Pi_{ \mathbf{M}}$ \label{AppD:ResSimu}}
The following figures show the performance of the nonparametric estimator $\hat{\Pi}_{ \mathbf{M}} $ of $\Pi_{ \mathbf{M}} $ for several scenarios beyond that shown in \autoref{FigA1}. The results are given over a range of $n$ and $r$ values as indicated in the figure captions.  Column 1: Boxplots of the difference between the row spaces spanned by $\mathbf{M}$ and $\mathbf{\hat{M}}$ as measured by $d(\mathbf{M},\mathbf{\hat{M}})$ when the true dimension of $\mathbf{M}$ is utilized to form $\mathbf{\hat{M}}$ (i.e., setting $\hat{r}=r$). Column 2: Boxplots of $d(\mathbf{M},\mathbf{\hat{M}})$ when using the proposed estimator $\hat{r}$ of the row space dimension in forming $\mathbf{\hat{M}}$.  Column 3:  An assessment of the estimate $\mathbf{\hat{D}}_k$ of $\mathbf{D}_k$, where the latter term is the average of the column-wise variances of $\mathbf{Y}$.  The difference is measured by $\Vert \mathbf{\hat{D}}_k - \mathbf{D}_k \Vert = \max\limits_{1\leq j\leq n} \vert \hat{\delta}_{kj}-\bar{\delta}_{kj} \vert $.

\begin{figure}[H]
\centering
\includegraphics[height=0.56\textheight]{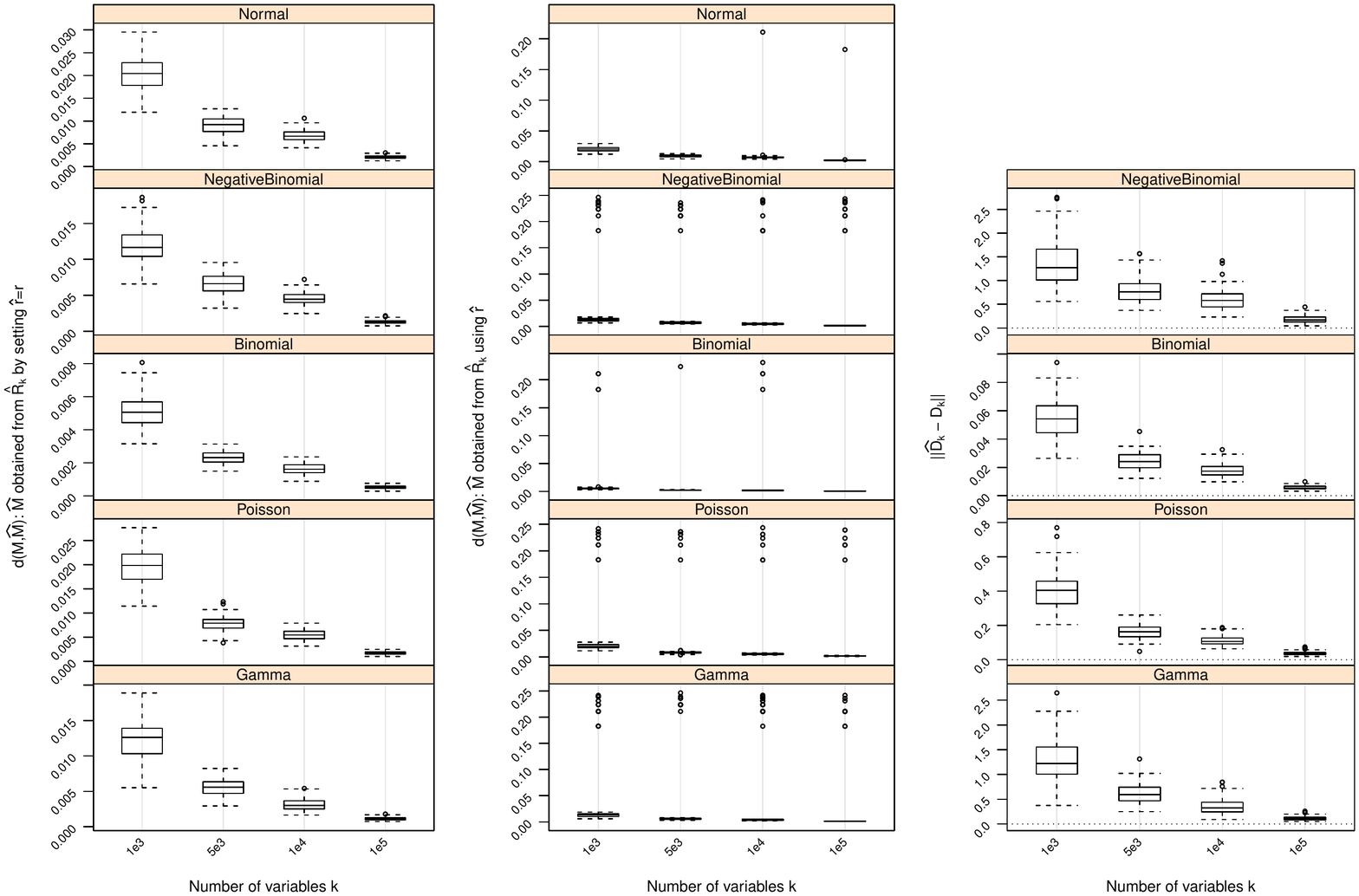}
\caption[Nonparametric Estimation of $\Pi_{ \mathbf{M}} $: $r=1$]{
Performance of $\hat{\Pi}_{ \mathbf{M}} $ when $n=15$ and $r=1$.
\label{FigA5}}
\end{figure}

\clearpage
\begin{figure}[H]
\centering
\includegraphics[height=0.56\textheight]{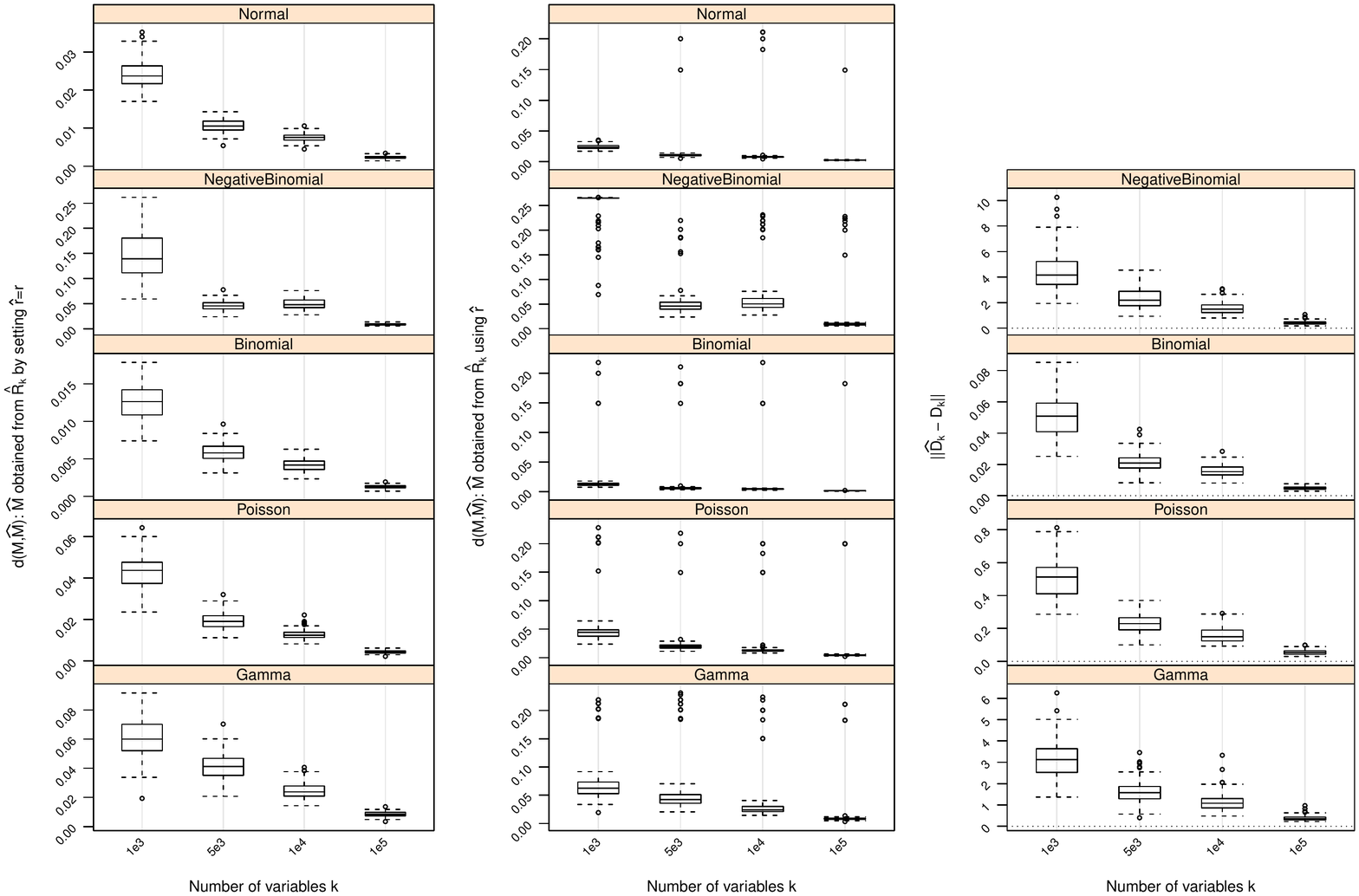}
\caption[Nonparametric Estimation of $\Pi_{ \mathbf{M}} $: $r=2$]{
Performance of $\hat{\Pi}_{ \mathbf{M}} $ when $n=15$ and $r=2$.
\label{FigA4}}
\end{figure}

\clearpage
\begin{figure}[H]
\centering
\includegraphics[height=0.56\textheight]{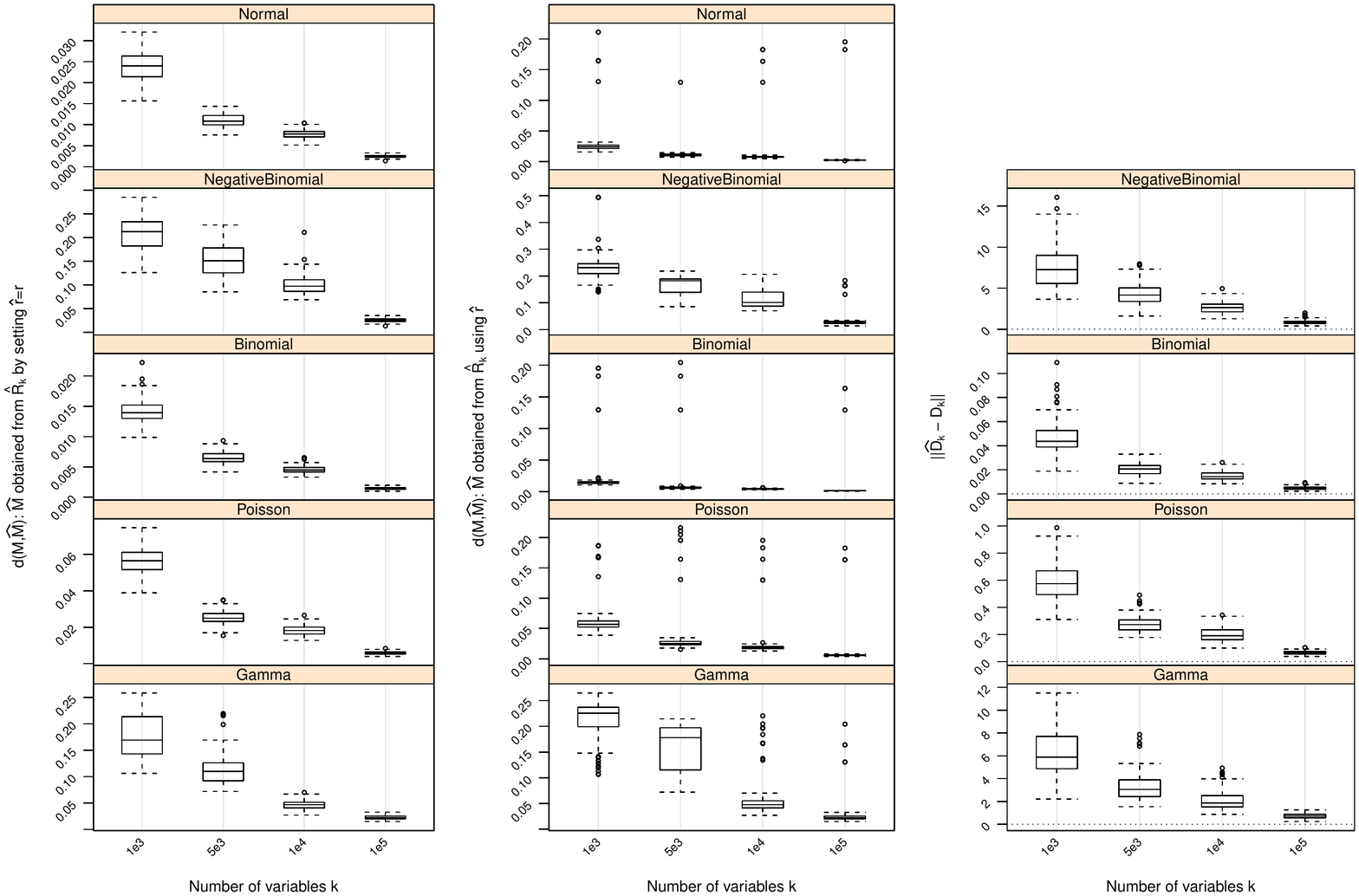}
\caption[Nonparametric Estimation of $\Pi_{ \mathbf{M}} $: $r=3$]{
Performance of $\hat{\Pi}_{ \mathbf{M}} $ when $n=15$ and $r=3$.
\label{FigA3}}
\end{figure}

\clearpage
\begin{figure}[H]
\centering
\includegraphics[height=0.56\textheight]{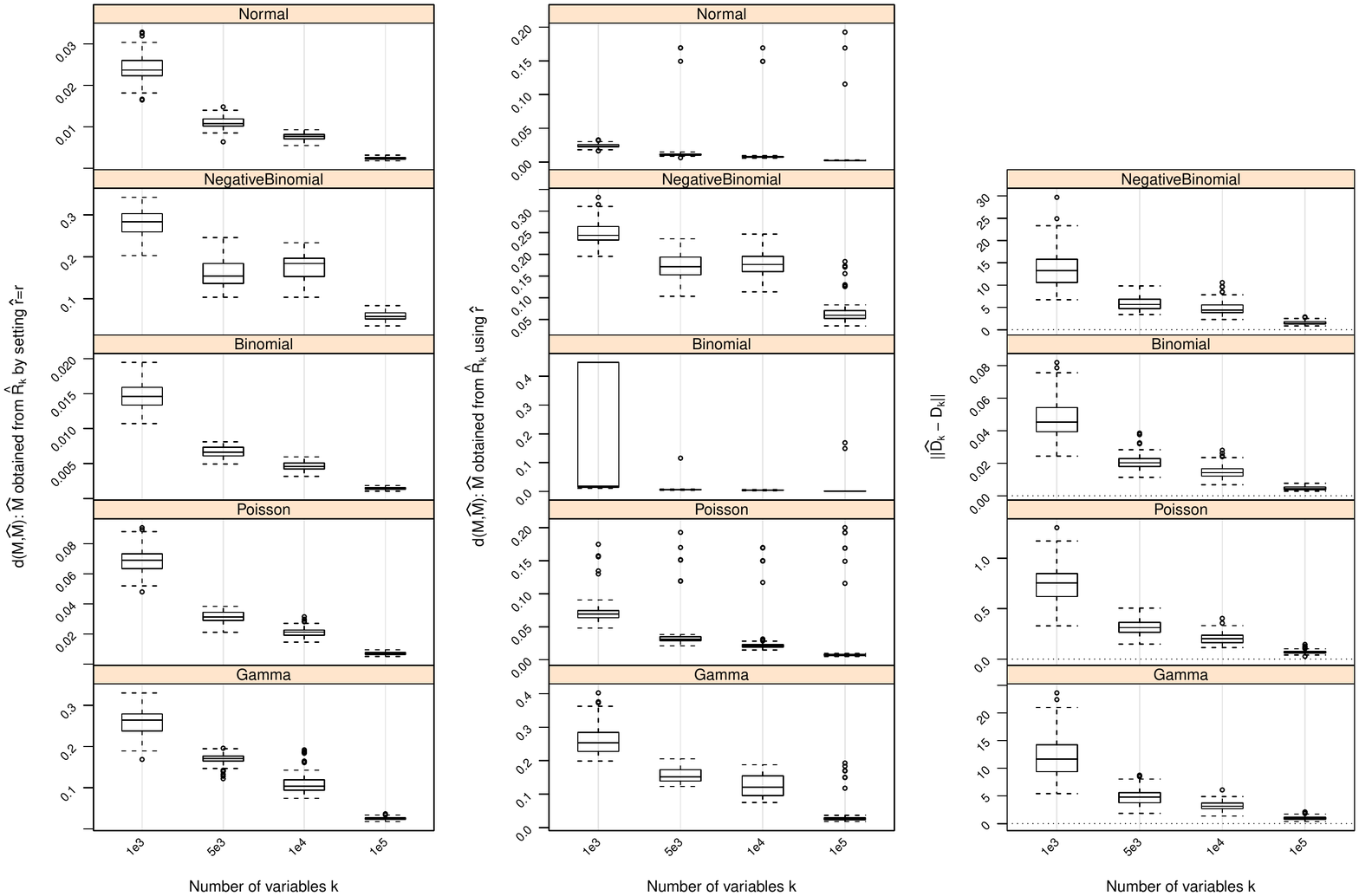}
\caption[Nonparametric Estimation of $\Pi_{ \mathbf{M}} $: $r=4$]{
Performance of $\hat{\Pi}_{ \mathbf{M}}$ when $n=15$ and $r=4$.
\label{FigA2}}
\end{figure}

\clearpage
\begin{figure}[H]
\centering
\includegraphics[height=0.56\textheight]{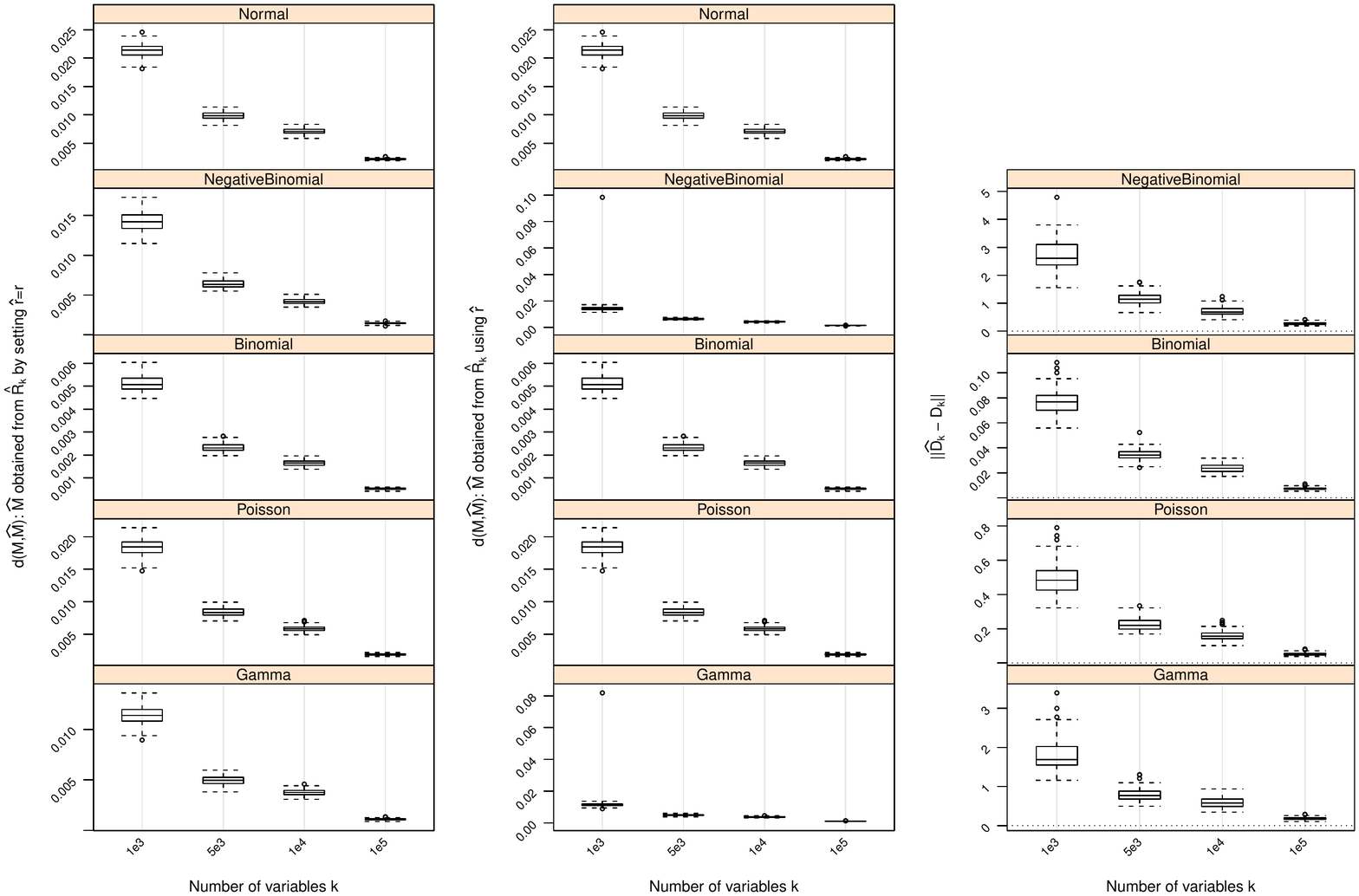}
\caption[Nonparametric Estimation of $\Pi_{ \mathbf{M}} $: $n=100, r=1$]{
Performance of $\hat{\Pi}_{ \mathbf{M}}$ when $n=100$ and $r=1$.
\label{FigB1}}
\end{figure}

\clearpage
\begin{figure}[H]
\centering
\includegraphics[height=0.56\textheight]{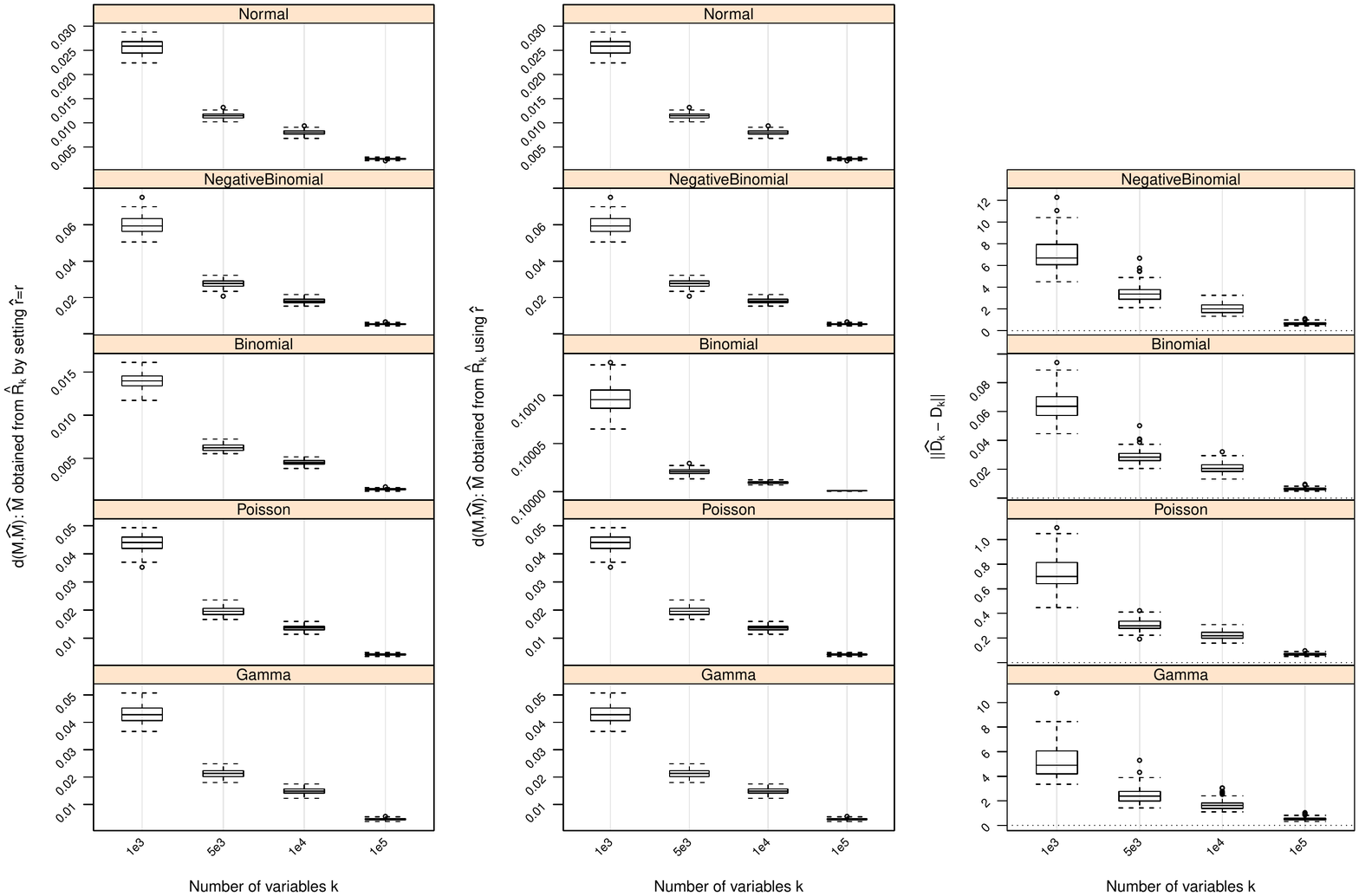}
\caption[Nonparametric Estimation of $\Pi_{ \mathbf{M}} $: $n=100, r=2$]{
Performance of $\hat{\Pi}_{ \mathbf{M}}$ when $n=100$ and $r=2$.
\label{FigB2}}
\end{figure}

\clearpage
\begin{figure}[H]
\centering
\includegraphics[height=0.56\textheight]{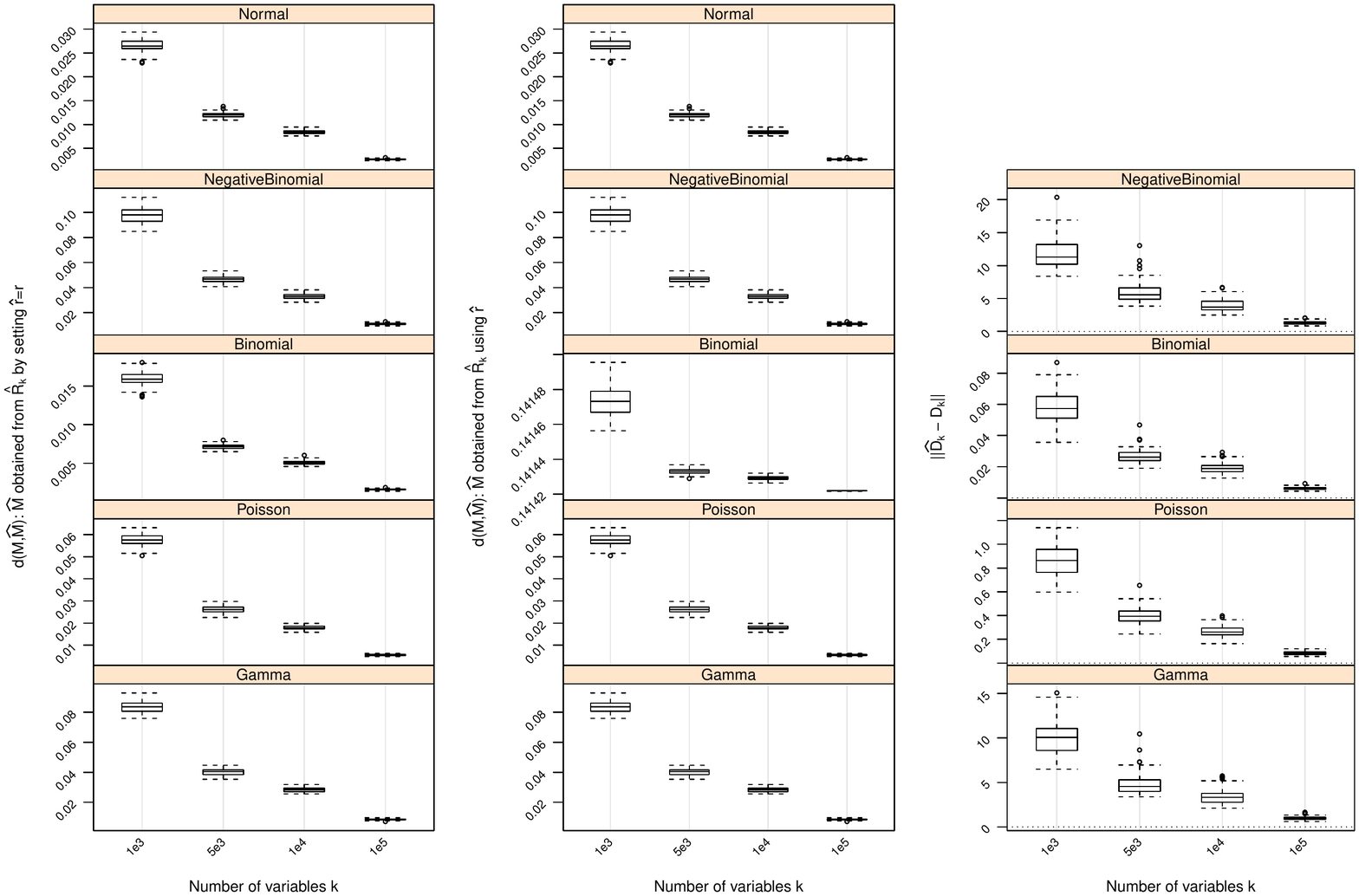}
\caption[Nonparametric Estimation of $\Pi_{ \mathbf{M}} $: $n=100, r=3$]{
Performance of $\hat{\Pi}_{ \mathbf{M}} $ when $n=100$ and $r=3$.
\label{FigB3}}
\end{figure}

\clearpage
\begin{figure}[H]
\centering
\includegraphics[height=0.56\textheight]{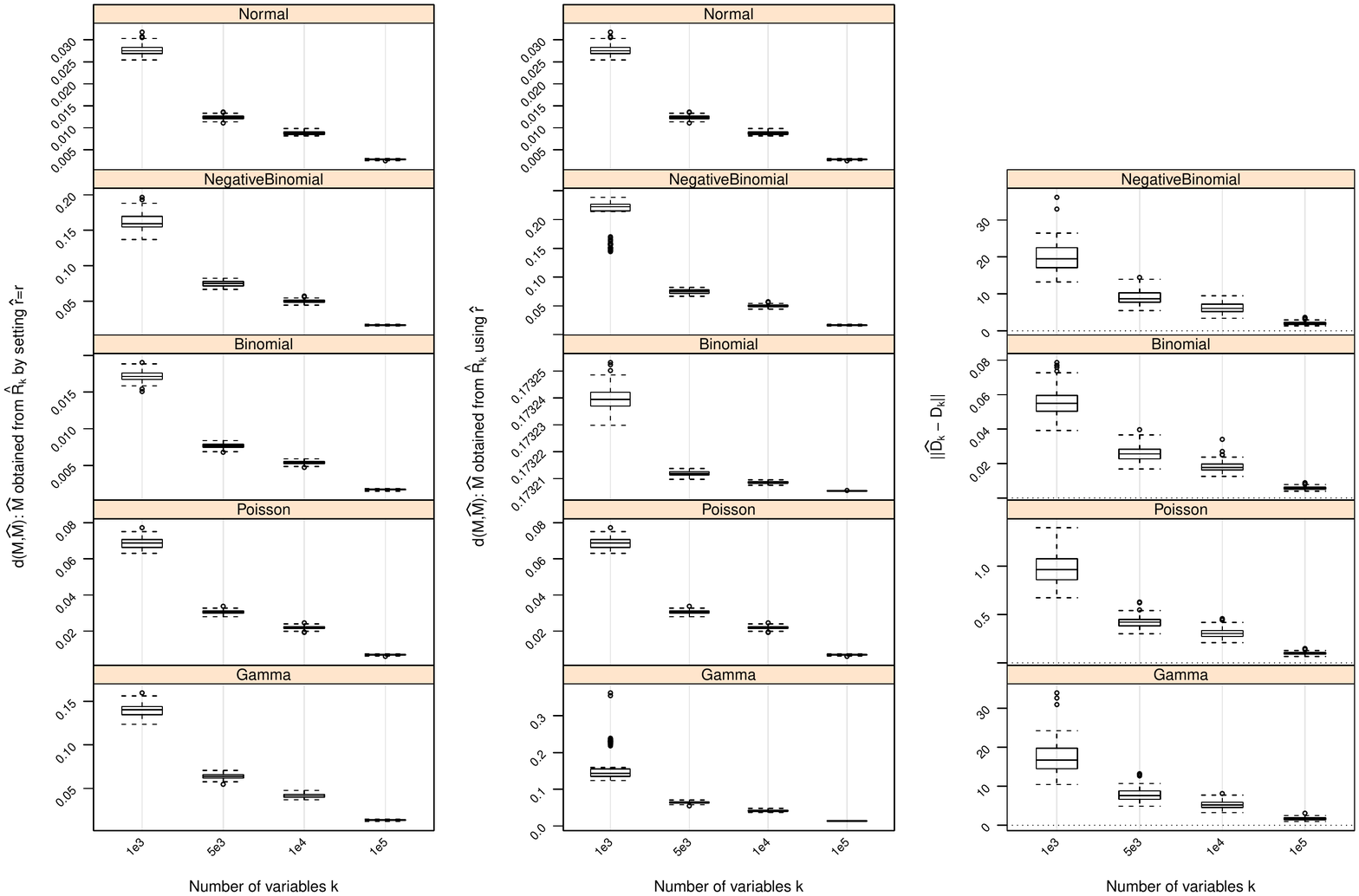}
\caption[Nonparametric Estimation of $\Pi_{ \mathbf{M}} $: $n=100, r=4$]{
Performance of $\hat{\Pi}_{ \mathbf{M}} $ when $n=100$ and $r=4$.
\label{FigB4}}
\end{figure}

\clearpage
\begin{figure}[H]
\centering
\includegraphics[height=0.56\textheight]{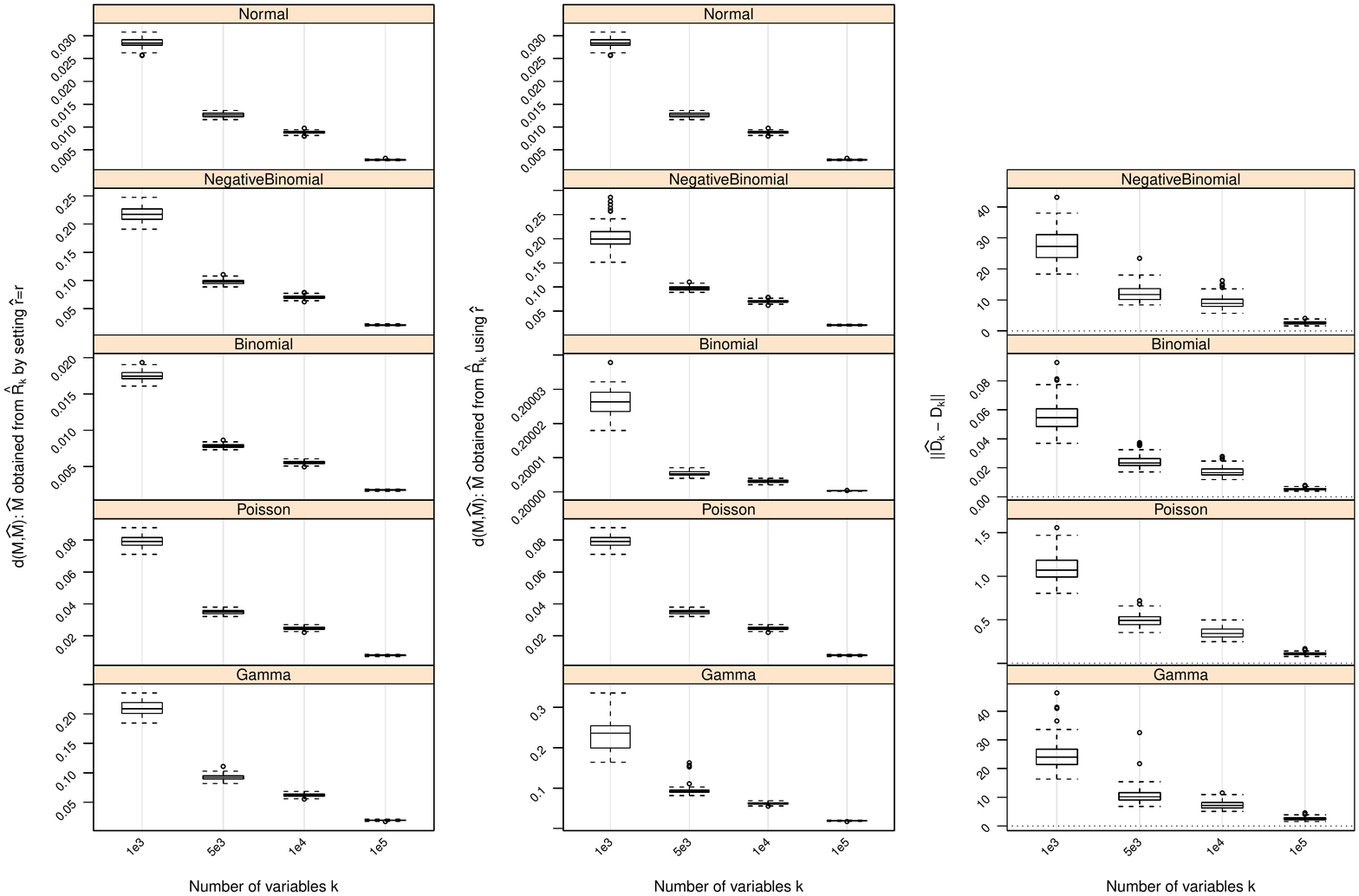}
\caption[Nonparametric Estimation of $\Pi_{ \mathbf{M}} $: $n=100, r=5$]{
Performance of $\hat{\Pi}_{ \mathbf{M}} $ when $n=100$ and $r=5$.
\label{FigB5}}
\end{figure}

\clearpage
\begin{figure}[H]
\centering
\includegraphics[height=0.56\textheight]{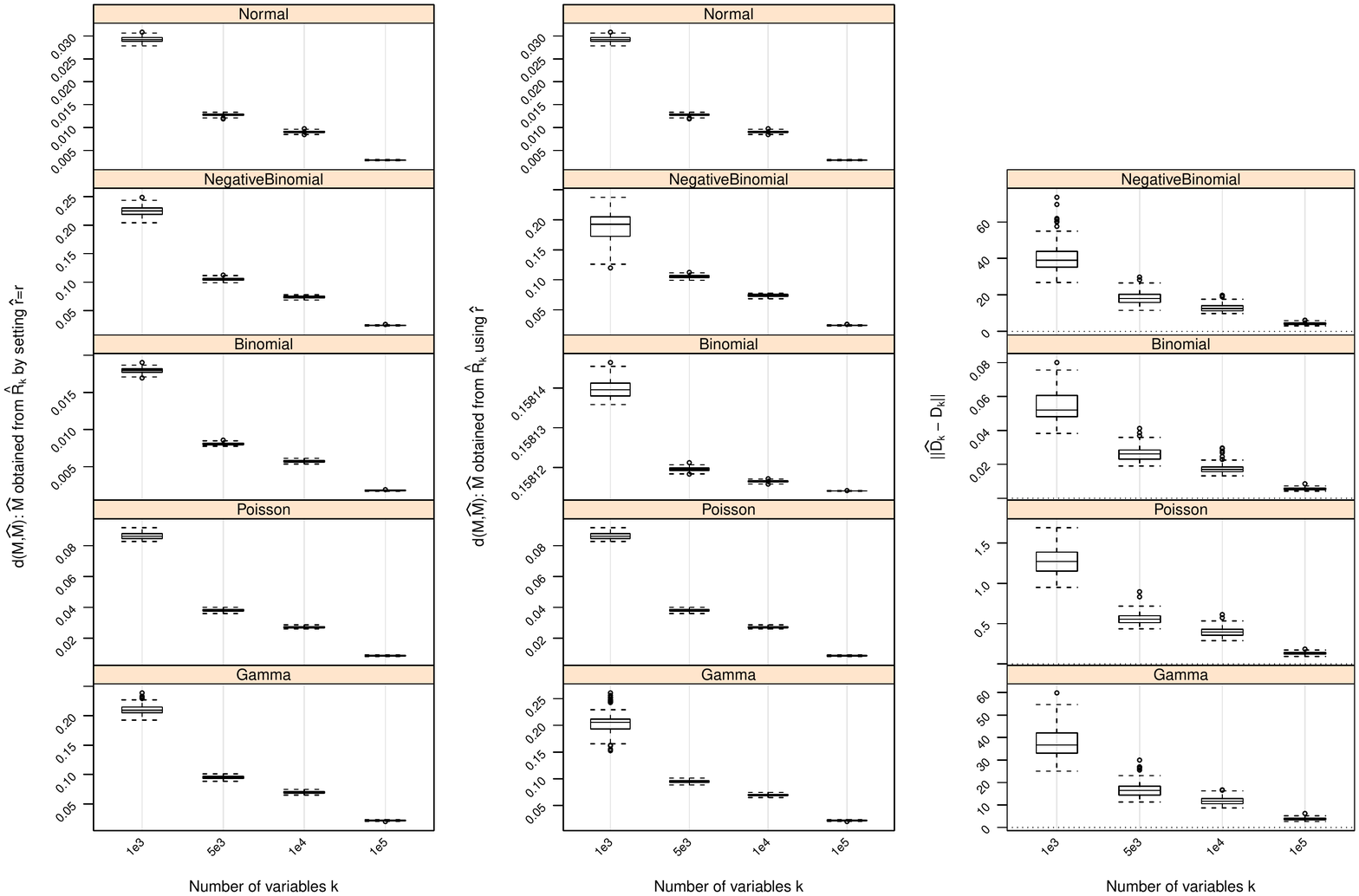}
\caption[Nonparametric Estimation of $\Pi_{ \mathbf{M}} $: $n=200, r=6$]{
Performance of $\hat{\Pi}_{ \mathbf{M}}$ when $n=200$ and $r=6$.
\label{FigC1}}
\end{figure}

\clearpage
\begin{figure}[H]
\centering
\includegraphics[height=0.56\textheight]{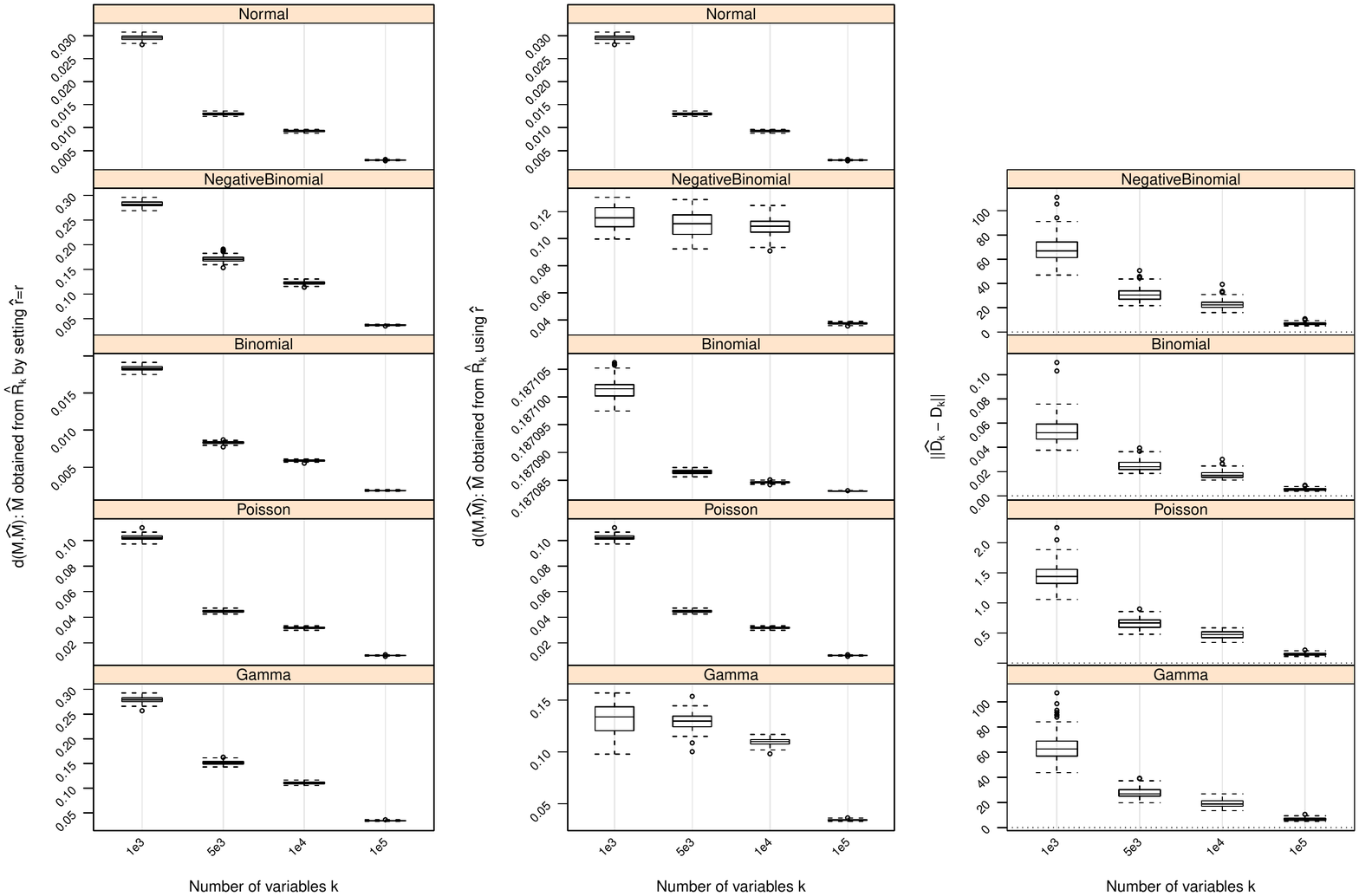}
\caption[Nonparametric Estimation of $\Pi_{ \mathbf{M}} $: $n=200, r=8$]{
Performance of $\hat{\Pi}_{ \mathbf{M}}$ when $n=200$ and $r=8$.
\label{FigC2}}
\end{figure}

\clearpage
\begin{figure}[H]
\centering
\includegraphics[height=0.56\textheight]{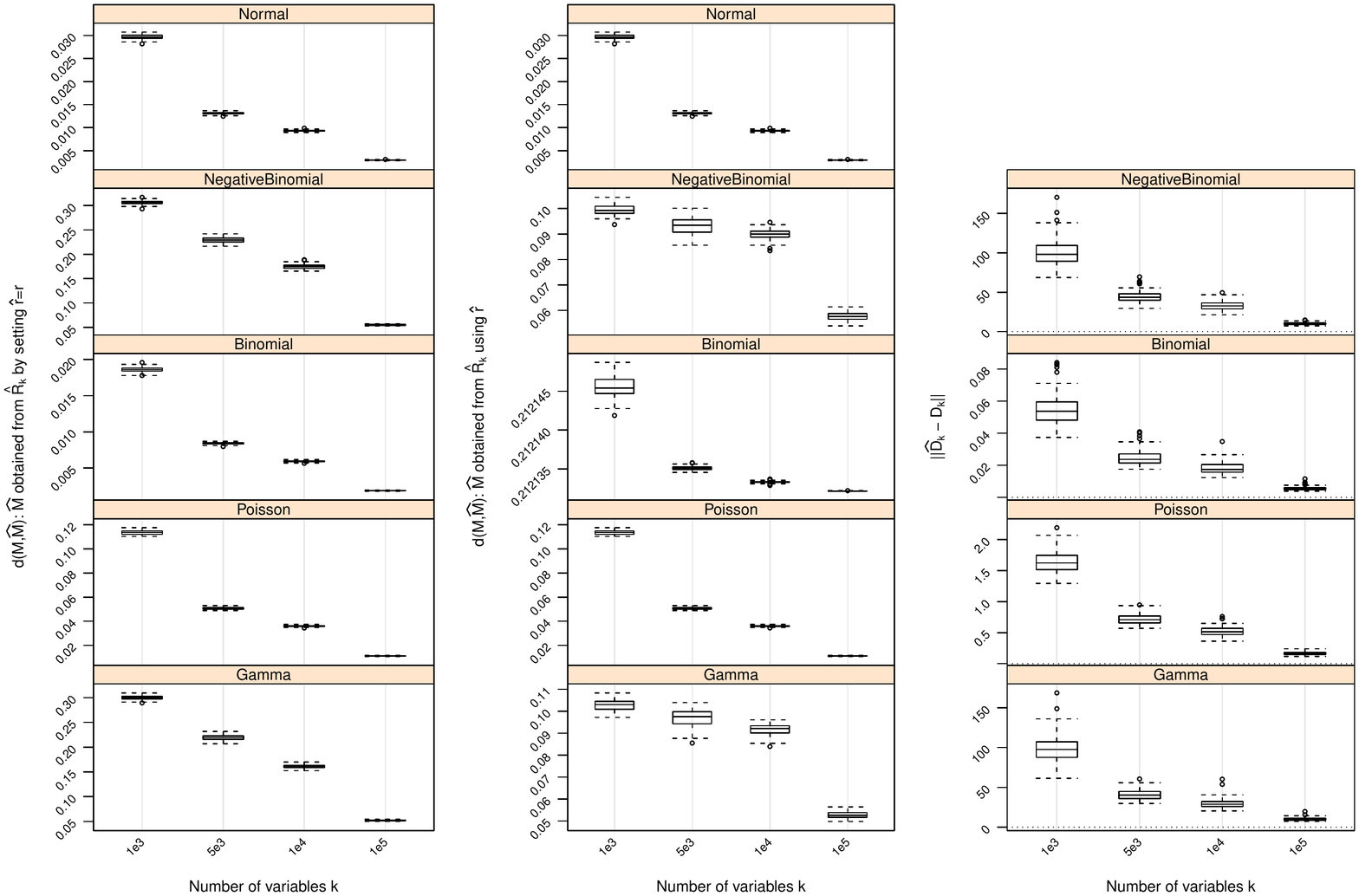}
\caption[Nonparametric Estimation of $\Pi_{ \mathbf{M}} $: $n=200, r=10$]{
Performance of $\hat{\Pi}_{ \mathbf{M}} $ when $n=200$ and $r=10$.
\label{FigC3}}
\end{figure}

\clearpage
\begin{figure}[H]
\centering
\includegraphics[height=0.56\textheight]{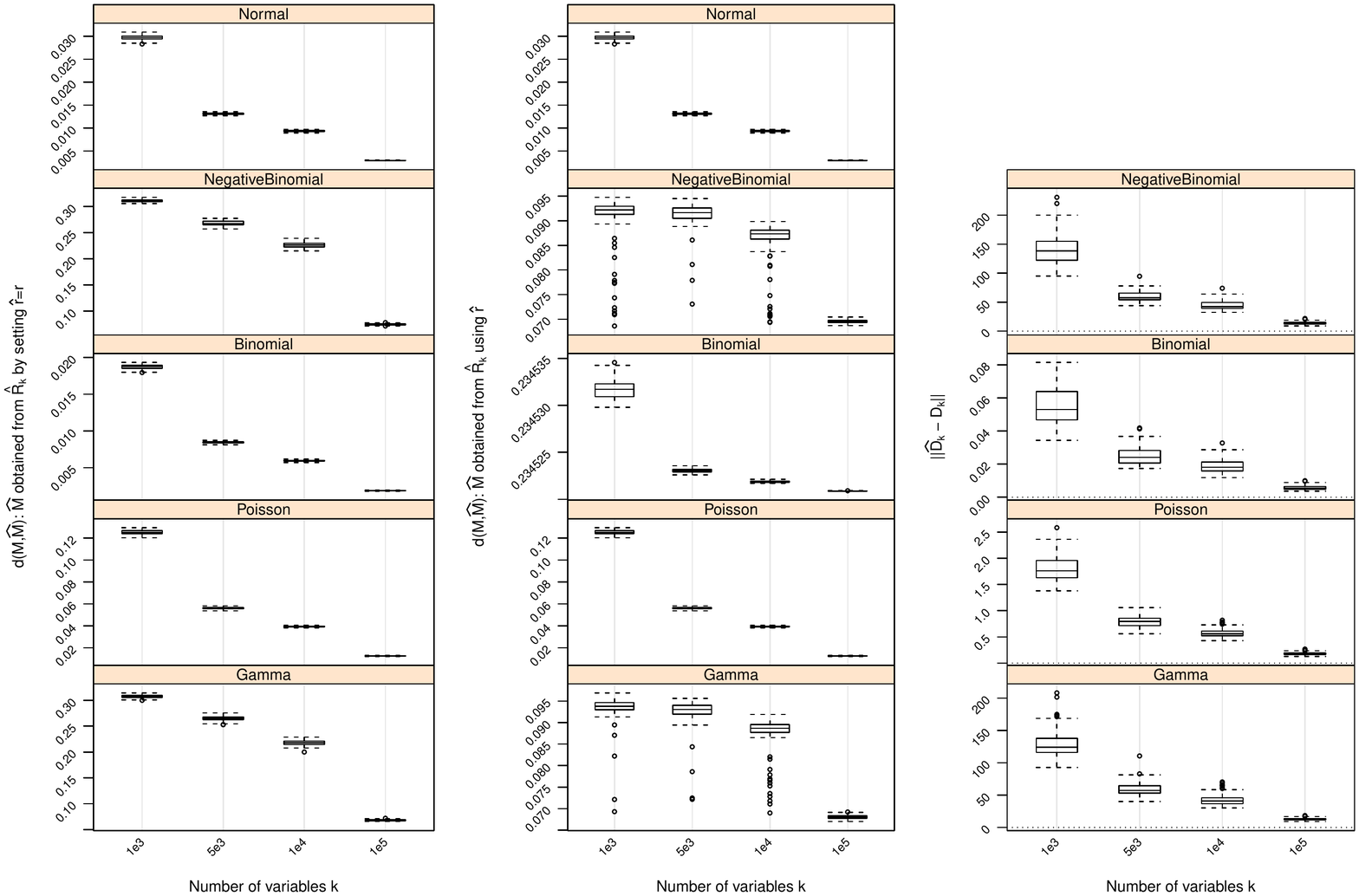}
\caption[Nonparametric Estimation of $\Pi_{ \mathbf{M}} $: $n=200, r=12$]{
Performance of $\hat{\Pi}_{ \mathbf{M}} $ when $n=200$ and $r=12$.
\label{FigC4}}
\end{figure}

\end{document}